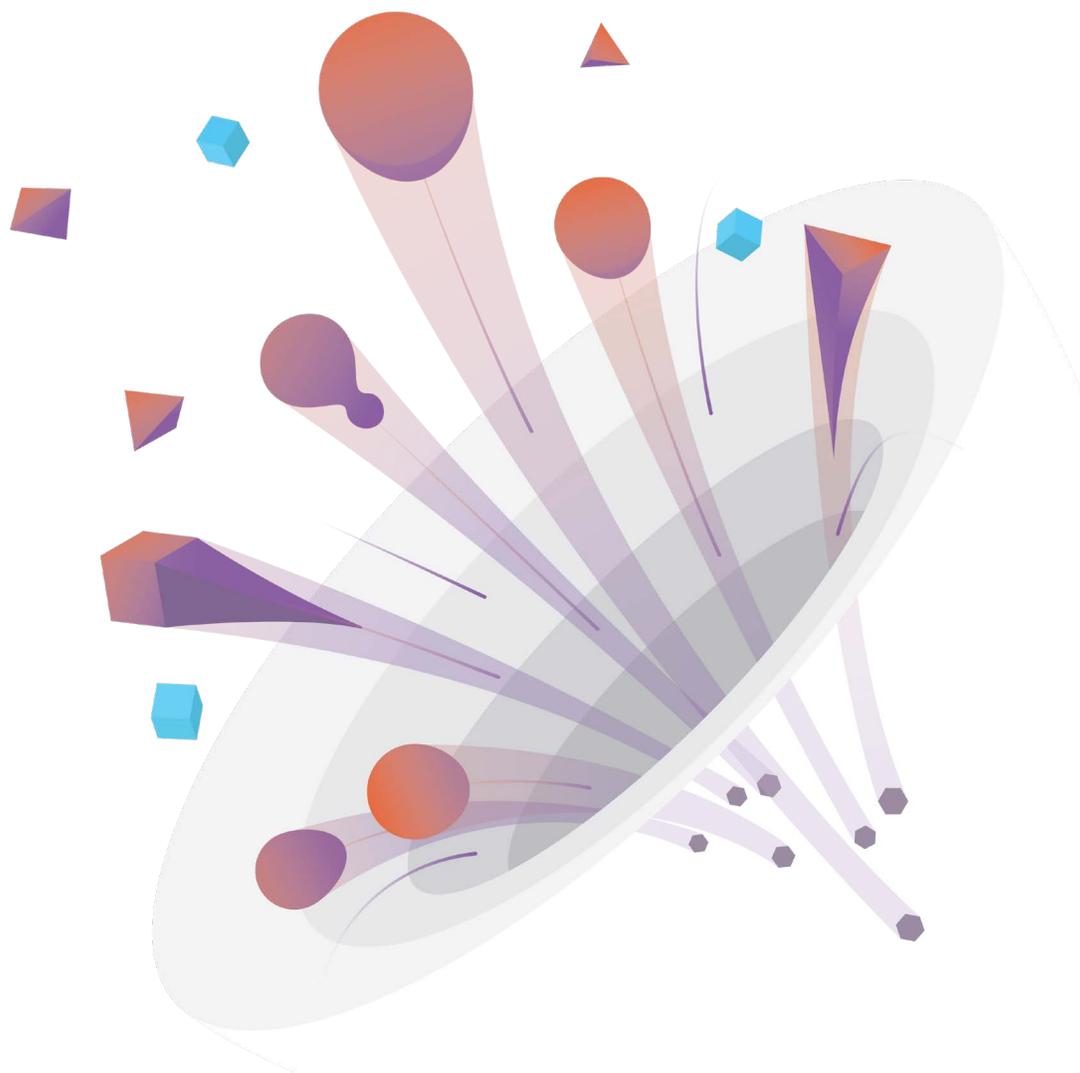

# Lost in Translation

## Large Language Models in Non-English Content Analysis

**Gabriel Nicholas**
**Aliya Bhatia**

**May 2023**

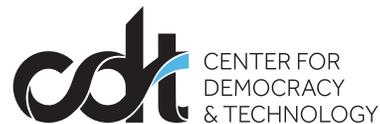

The Center for Democracy & Technology (CDT) is the leading nonpartisan, nonprofit organization fighting to advance civil rights and civil liberties in the digital age. We shape technology policy, governance, and design with a focus on equity and democratic values. Established in 1996, CDT has been a trusted advocate for digital rights since the earliest days of the internet. The organization is headquartered in Washington, D.C., and has a Europe Office in Brussels, Belgium.

---

**GABRIEL NICHOLAS**

Research Fellow at the Center for Democracy & Technology.

**ALIYA BHATIA**

Policy Analyst, Free Expression Project at the Center for Democracy & Technology.

cdt | **Research**

# Lost in Translation

## Large Language Models in Non-English Content Analysis

### Gabriel Nicholas and Aliya Bhatia

**WITH CONTRIBUTIONS BY**

Samir Jain, Mallory Knodel, Emma Llansó, Michal Luria, Nathalie Maréchal, Dhanaraj Thakur, and Caitlin Vogus.

**ACKNOWLEDGMENTS**

We thank Pratik Joshi, Sebastin Santy, and Aniket Kesari for their invaluable feedback on the technical aspects of this report. We also thank Jacqueline Rowe, Damini Satija, and Ángel Díaz for their insightful comments and suggestions. The translation of our executive summary is made possible by Global Voices Translations and with the help of Iverna McGowan, Maria Villamar, Ophélie Stockhem, and Tomás Pomar. All views in this report are those of CDT.

This work is made possible through a grant from the John S. and James L. Knight Foundation.

**Suggested Citation:** Nicholas, G. and Bhatia, A. (2023) Lost in Translation: Large Language Models in Non-English Content Analysis. Center for Democracy & Technology. https://cdt.org/insights/lost-in-translation-large-language-models-in-non-english-content-analysis/

References in this report include original links and links archived and shortened by the Perma.cc service. The Perma.cc links also contain information on the date of retrieval and archive.

# Contents

# Executive Summary

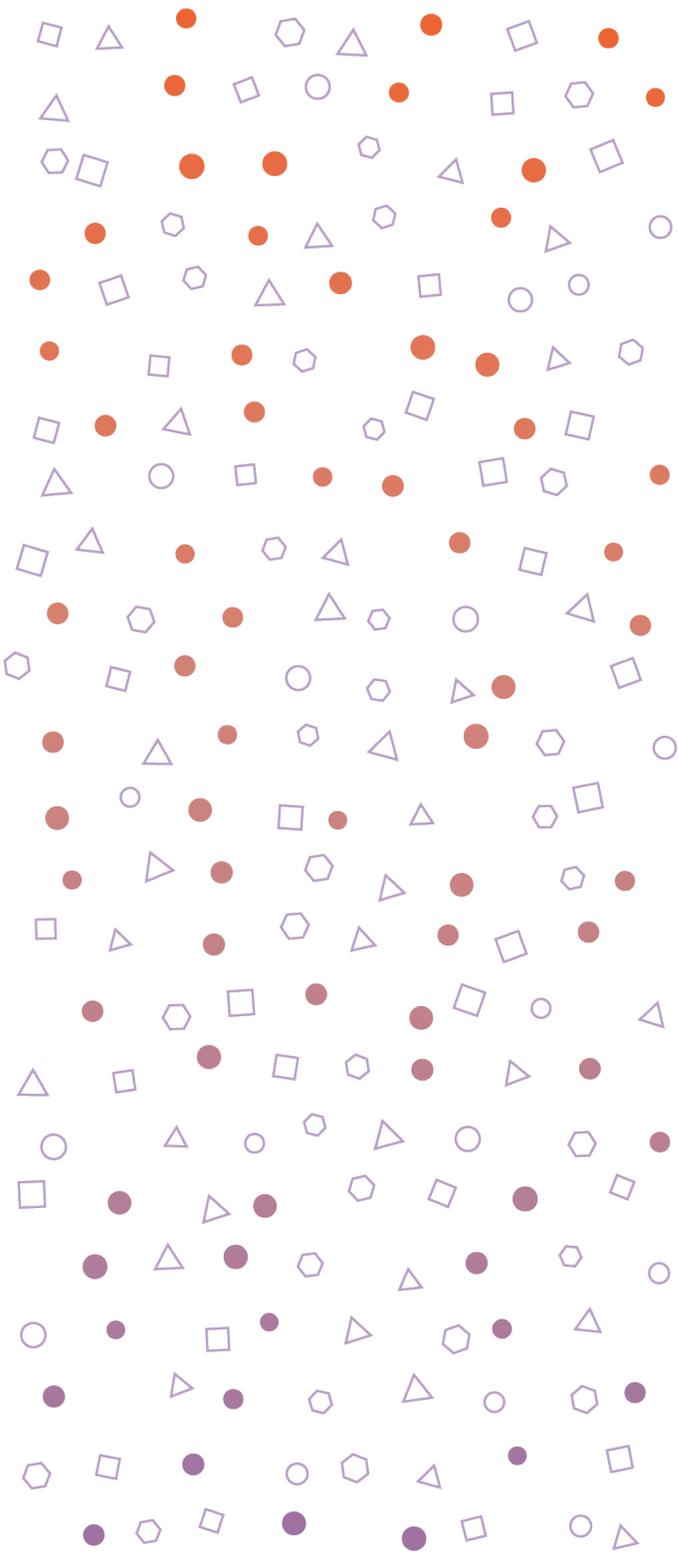

The internet is the primary source of information, economic opportunity, and community for many around the world. However, the automated systems that increasingly mediate our interactions online — such as chatbots, content moderation systems, and search engines — are primarily designed for and work far more effectively in English than in the world's other 7,000 languages.

In recent years, large language models have become the dominant approach for building AI systems to analyze and generate language online, but again, they have been built primarily for the English language. A large language model (e.g., Open AI's GPT-4, Meta's LLaMa, Google's PaLM) is a machine learning algorithm that scans enormous volumes of text to learn which words and sentences frequently appear near one another and in what context. Large language models can be adapted to perform a wide range of tasks across different domains. They are most known for being used to build chatbots, such as ChatGPT, but researchers and technology companies also use them for content analysis tasks, such as sentiment analysis, text summarization, and hate speech detection. Google, Meta, Microsoft, and other companies have already incorporated large language models into their core product functions, such as content moderation and search. Other vendors soon may incorporate them into automated decision-making systems, such as resume scanners.

Recently though, researchers and technology companies have attempted to extend the capabilities of large language models into languages other than English by building what are called *multilingual language models*. Instead of being trained on text from only one language, multilingual language models are trained on text from dozens or hundreds of languages at once. Researchers posit that multilingual language models infer connections between languages, allowing them to apply word associations and underlying grammatical rules learned from languages with more text data available to train on (in particular English) to those with less. In some applications, multilingual language models outperform models trained on only one language — for instance, a model trained on lots of text from lots of languages, including Hindi, might perform better in Hindi contexts than a model just trained on Hindi text.

Multilingual language models give technology companies a way to scale their AI systems to many languages at once, and some have already begun to integrate them into their products. Online service providers in particular have deployed multilingual language models to moderate content: Meta uses a multilingual language model to detect harmful content on its platforms in over 100 languages; Alphabet's Perspective API uses one to detect toxic content in eighteen different languages; Bumble uses one to detect and take action on unwanted sexual messages around the world.

Multilingual language models allow technologists to attempt to build models in languages for which they otherwise might not have enough digitized text. Languages vary widely in *resourcedness*, or the volume, quality, and diversity of text data they have available to train language models on. English is the highest resourced language by multiple orders of magnitude, but Spanish, Chinese, German, and a handful of other languages are sufficiently high resource enough to build language models in. Medium resource languages, with fewer but still high-quality data sets, such as Russian, Hebrew, and Vietnamese, and low resource languages, with almost no training data sets, such as Amharic, Cherokee, and Haitian Creole, have too little text for training their own large language models. Language data in low resource languages is also often of particularly poor quality: either it is mistranslated or even nonsensical language scraped from the internet, or is limited to sources with narrow domains, such as religious texts and Wikipedia. This gap in data availability between languages is known as the *resourcedness gap*.

Multilingual language models are designed to address these gaps in data availability by inferring semantic and grammatical connections between higher- and lower-resource languages, allowing the former to bootstrap the latter. However, this architecture raises its own concerns. Multilingual language models are still usually trained disproportionately on English language text and thus end up transferring values and assumptions encoded in English into other language contexts where they may not belong. For example, a multilingual model might associate the word "dove" in all languages with "peace" even though the Basque word for dove ("uso") can be an insult. The disparity in available data also means multilingual language models work far better in higher resource languages and languages similar to them than lower resource ones. Model developers will sometimes try to fill in these gaps with machine-translated text, but translation errors may further compound language misrepresentation. And when multilingual language models do fail, their unintuitive connections between languages can make those problems harder to identify, diagnose, and fix.

Large language models' general use in content analysis raises further concerns. Computational linguists argue that large language models are limited in their capacity to analyze forms of expression not included in their training data, meaning they may struggle to perform in new contexts. They may also reproduce any biases present in their training data. Often, this text is scraped from the internet, meaning that large language models may encode and reinforce dominant views expressed online.

Companies, researchers, and governments each have a role to play in protecting the public from the potential dangers of multilingual language model content analysis systems. To ensure better public accountability, companies that deploy large language models should always be transparent about how they use them and in which languages. Companies should deploy language models with narrow remits and adequate channels for human review.

Researchers and research funders meanwhile should invest in efforts to improve the use and performance of language models in languages other than English, in particular, to reduce failures that disparately impact speakers of lower-resourced languages. The best way to do this is by supporting language-specific research communities, who can promote the virtuous cycle of collecting data, curating datasets, training language models, publishing, and building applications. Local language speakers and context experts need to be part of each step of this process and also be curating the data and assessing the language models deployed by large, global online services.

Finally, governments need to be careful about how they use or encourage the use of large language models. Large language models should never power systems used to make high-stakes decisions without oversight, such as decisions about immigration status or healthcare, nor should governments mandate or inadvertently require by law the use of large language model-powered systems to moderate content from online services. Instead, governments should convene different stakeholders to align on what norms and guardrails should be around developing and deploying large language models.

Large language models in general and multilingual language models in particular have the potential to create new economic opportunities and improve the web for all. However, mis- or over-application of these technologies poses real threats to individuals' rights, such as undermining their right to free expression by inaccurately taking down a person's post on social media or their right to be free of discrimination by misinterpreting an individual's job or visa application. Multilingual language models specifically can inadvertently further entrench the Anglocentrism they are intended to address. In light of these limitations, technology companies, researchers, and governments must consider potential human and civil rights risks when studying, procuring, developing, or using multilingual language models to power systems, in particular when they are used to make critical information available or play a role in decisions affecting people's access to economic opportunities, liberty, or other important interests or rights.

# Introduction

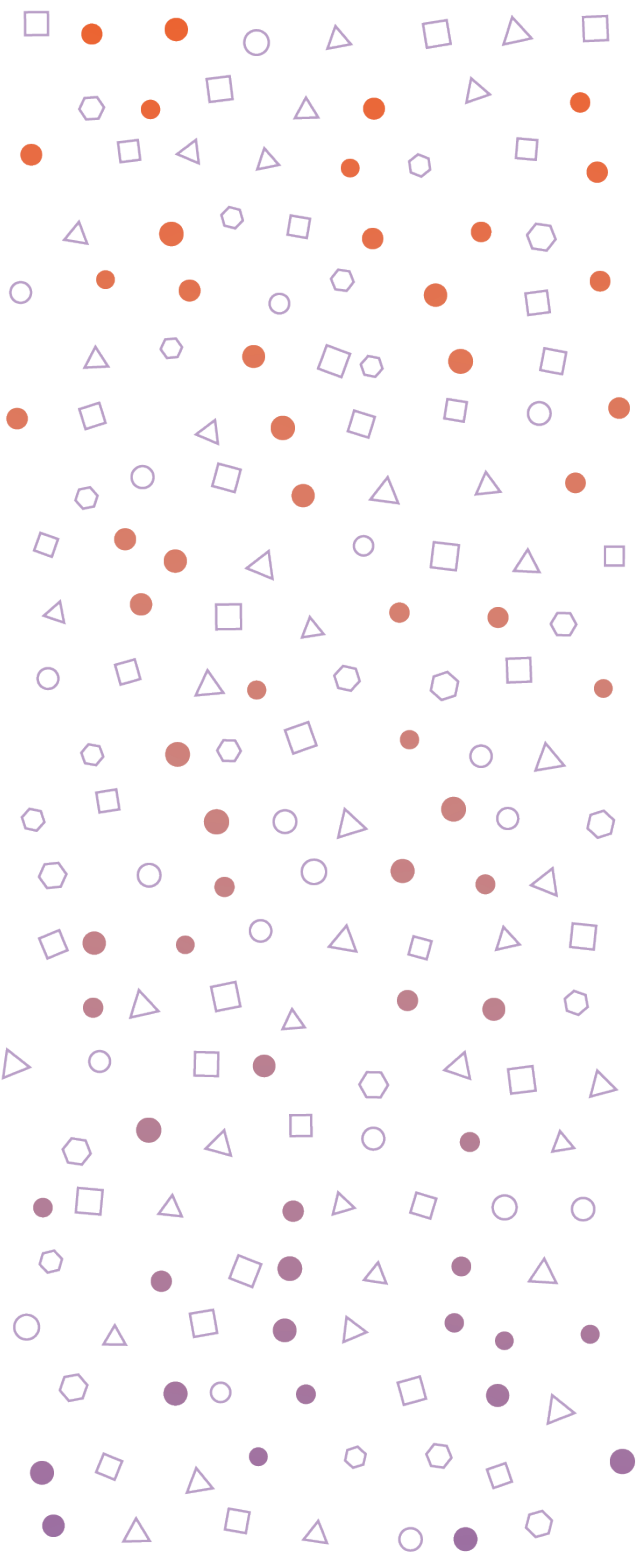

Despite the modern internet's power to mobilize and connect people around the world, the web still does not reflect the linguistic diversity of its users. In particular, the automated systems that increasingly mediate our interactions online — such as chatbots, search engines, and content moderation systems — are built using and perform far better on English-language text than the world's other 7,000 languages (Kornai, 2013; Sengupta, 2022). Individuals speaking languages other than English face barriers to expressing themselves freely online and may face greater challenges when it comes to accessing critical information, public services, and even asylum and safety (Torbati, 2019).

In the last few years, however, there have been rapid advancements in developing machine learning tools that can analyze content in a wide variety of languages and across different domains. *Large language models*, machine learning tools trained on enormous amounts of text to recognize patterns in language, power many of these systems. Large language models already underlie translation apps, search autocomplete, and chatbots such as ChatGPT. They are known for being adaptable to many different language tasks, and today, researchers and technologists are constantly on the lookout for new applications and contexts in which to deploy them. Since the late 2010s, major U.S.-based technology companies have mostly invested in building large language models that work primarily for English, such as Open AI's GPT-4, Meta's LLaMa, and Google's PaLM.

Recently, companies and researchers have begun building and researching *multilingual language models*, large language models trained on text data from several different languages at once. Meta's XLM-RoBERTa (XLM-R) for instance is trained on text from 100 languages (Meta AI, 2019) at once. Google's mBERT, a multilingual version of its popular BERT model, is trained on 104 languages. Researchers claim that these models extend the multifaceted capabilities of large language models to languages other than English, even to languages for which there is little or no text data for the model to learn from (Artetxe & Schwenk, 2019; Wu & Dredze, 2019).

Technology companies have their own interests in improving how well large language models work in different languages. Some may want to make their products available in multiple languages to gain a competitive edge in emerging and populous markets. Online services that host user-generated content may especially be interested in using multilingual language models to detect and take action on hate speech, disinformation, and other content that violates their policies or the law (Dulhanty et al., 2019). This is top of mind for services after facing criticism for not taking more aggressive action against content that incited violence and genocide in Ethiopia, Nigeria, and Myanmar, among others. Services have begun to deploy multilingual language models into their content moderation systems: Meta claims their XLM-R model can detect harmful content in all 100 languages it is trained on (Meta AI, 2021); Alphabet's Perspective API uses a large language model to detect toxic content in eighteen different languages (Lees et al., 2022); Bumble uses one to detect rude and abusive messages in at least fifteen languages (Belloni, 2021). Technology companies are also repurposing these models to make health care information available and soon may reach into other domains as well (Lunden, 2023).

In the future, governments could also seek to use automated systems built using large language models to make information available, answer questions in languages spoken by their constituents (in the form of chatbots), or, more dangerously, analyze information to make critical decisions such as benefits allocation or refugee status determinations (Kinchin & Mougouei, 2022).

Still, studies show that even multilingual language models struggle to deal with the wide disparities between different languages in how much text data they have available to train and test language models. English has, by multiple orders of magnitude, more text data available than any other language and commands most of the attention of the natural language processing research community. The abundance of English language data stems from its position as the official or de facto language of international business, politics, and media, itself a legacy of British colonialism and American neocolonialism and the subsequent erasure of regional and indigenous languages. American technology companies have further entrenched English as the predominant language of the internet by rolling out early standards, coding languages, and social media platforms in English long before other languages.

The hegemony of English data means that most large language models, even multilingual ones, are built predominantly using Standard English language text and work best in Standard English language contexts. Spanish, Chinese, Arabic, and a few other "high resource" languages also have significant amounts of text data available, but many "medium resource" languages, such as Hindi and Portuguese, and "low resource" languages, such as Haitian Creole and Swahili, have hardly any data available at all, and multilingual language models perform much worse in those languages. This skewed emphasis fails to reflect the diversity of languages spoken by the world's internet users and further perpetuates the dominance of the English language.

Despite being deployed in real-world systems, multilingual language models have largely been absent from public discourse, particularly about digital rights and public policy, and have instead been relegated to computer science academia and tech company public relations. This paper seeks to address this gap by offering several resources to bolster policy discussions. Part I provides a simple technical explanation of how large language models work in general, why there is a gap in available data between English and other languages, and how multilingual language models attempt to bridge that gap. Part II accounts for the challenges of doing content analysis with large language models in general and multilingual language models in particular, namely:

1.  Multilingual language models often rely on machine-translated text that can contain errors or terms native language speakers don't actually use.

2.  When multilingual language models fail, their problems are hard to identify, diagnose, and fix.

3.  Multilingual language models do not and cannot work equally well in all languages.

4.  Multilingual language models fail to account for the contexts of local language speakers.

Finally, Part III provides recommendations for companies, researchers, and policymakers to keep in mind when considering studying, developing, and deploying large and multilingual language models to do content analysis. These recommendations offer guidance concerning when large language models should or should not be deployed, how to improve their performance in non-English languages, and how to ensure better accountability and transparency to local language stakeholders.

Before proceeding, two notes on the terminology used in this primer. First, this paper focuses specifically on one category of applications for large language models: *content analysis*, or, the inference and extraction of information, themes, and concepts from text. The Center for Democracy & Technology (CDT) has written many times about the limitations of automated content analysis systems (Duarte et al., 2017; Shenkman et al., 2021) and the civil liberty risks they can pose, particularly in areas such as content moderation, student activity monitoring, hiring and more (Grant-Chapman et al., 2021; Nicholas, 2022; Vallee & Duarte, 2019). Large language models are already deeply integrated into many of these technical systems, particularly content moderation, and will soon become part of many more. Public discourse about large language models has so far disproportionately focused on text generation, an important but not the only one. Many of the shortcomings of large language models presented in this report also apply to text generation. As such, this report can be read as a primer on some of the limits of generative AI systems as well. However, we choose to focus on content analysis for this report because of the potential dangers associated with using these models to host and make information available and the impacts on free expression rights.

Second, this paper focuses on how multilingual language models perform in languages other than English. We use the shorthand of "non-English languages" for easy reading and because it is the terminology used in the machine learning and policy literature. We recognize the irony that this term centers the English language and misleadingly implies all other languages are a monolith. Where possible, we elaborate upon the types of languages we are writing about and make distinct references to specific languages and cultural contexts that will elude models trained primarily in English. In some instances, we think the term "non-English" captures the sheer Anglocentrism of many of these models well by articulating the limited scope in which they are trained and tested.

# I. Background

## A. How Large Language Models Work

Natural language processing (NLP) is a subfield of artificial intelligence and linguistics concerned with building computer systems that can process and analyze language. NLP underlies many technologies we encounter every day — spellcheck, voice assistants like Siri or Alexa, resume scanners, language translators, and automated hate speech detection tools, to name a few. Until only a few years ago, when technologists wanted to teach a computer to perform a given NLP task, they would build a system specifically tailored to that task. To create a spam detection system for instance, a technologist might gather many emails, mark which ones are and are not spam, use some of those emails to train an algorithm and use others to test how well that algorithm works.

Today though, the field has fundamentally reoriented itself around repurposing *large language models* to solve nearly every problem. A *language model* is a mathematical function trained to solve a text prediction task like the following, "Given a sequence of words, predict what word will likely come next." For example, a language model might be given the phrase "I was a bad student, I used to skip ____," and generate as an output that there is a high percent chance the missing word is "class," a low percent it is "rope," and a near zero percent it is "clamoring."

The distribution of language that the model learns in the process can easily be repurposed to many different language tasks. The most often discussed application is text generation: conversational agents like ChatGPT can repurpose this text prediction task to answer questions, summarize text, and generate overall "human"-sounding speech. However, chatbots are just one application of large language models. Once a large language model is built, it can be further trained on a smaller dataset to improve its performance in a specific task, a process called *fine-tuning*. Today, for example, a developer building a spam detection system might take a general large language model already built by someone else — say Google's BERT — and fine-tune it to the specific task of spam detection using a handful of emails already labeled spam or not spam. By building it on top of a language model, the spam detection system will do a better job of detecting spam that doesn't perfectly match the language available in the email dataset.

Language models are not new. Computational linguists have used statistical models to try to infer rules about language since the 1980s (Nadkarni et al., 2011) and have used "neural networks" (an algorithm loosely modeled on how neurons connect in the brain) to do so since the early 2010s (Mikolov et al., 2013). What is new though is their largeness. Early language models could not be trained on as much data, since they had to read text in sequence, a process that could not be sped up by using more computing power. These early language models struggled to analyze words within the broader context of a sentence or document: for instance, one fine-tuned to detect suicidal ideation might have difficulty distinguishing between expressions of self-harm ("I just wish I was dead") and humor ("omg I'm dead"). But in 2017, Google researchers released a paper on a new architecture called *transformers*, which allowed language models to train on lots of data at the same time, in parallel rather than in sequence (Vaswani et al., 2017). These transformer-based language models could ingest so much data simultaneously that they could learn associations between entire sequences of words, not just individual words. Instead of being shown just {"dead"}, the model would see a word in its entire context, {"dead", ["omg", "I'm", "_____"]}, thus creating a much richer representation of language. Today, the only limit on the size of a language model — how much data it ingests and how many connections it makes between different sequences of words (i.e. *parameters*) — is how much data one can find and how much developers are willing to spend on processing power.

The output a language model produces is called a representation space, a map of the sequences of words that commonly appear near one another in the training text. For example, the phrases "It's so cold outside!" and "I better wear a jacket" may be near one another in a language model's representation space, since those sentences often appear close to one another in writing. This kind of proximity can lead to language models inferring patterns within language that can then help them conduct tasks that it is not explicitly trained in. In this case, sentences about cold weather being mapped near each other mean the large language model could be trained to detect whether a given phrase is about temperature.

With enough data, a large language model may have such a rich and multifaceted representation of a language that it can learn to do new tasks with only a few, or even zero examples to fine-tune on. For instance, the spam detection system described earlier could be built with little to no spam to fine-tune on. This capability is called "few-shot" or "zero-shot learning" and is one of the greatest promises of large language models, so much so that the original GPT-3 white paper is entitled "Language Models are Few-Shot Learners" (Brown et al., 2020).

Importantly though, large language models only learn the distribution of language, not its meaning (Bender & Koller, 2020). In the previous "cold" example, the model has not learned that when one is cold, one puts on a jacket or anything about the deeper meanings of "cold" and "jacket," only that the words often appear near one another. If one of the documents a large language model trains on is a humorous blogpost about the best shorts to wear in cold temperatures, the model could just as easily learn that "shorts" and "cold" are related. Similarly, if a model is trained only on very formal language data, it may never learn that "nippy" or "brick" (New York City slang) can refer to cold as well.

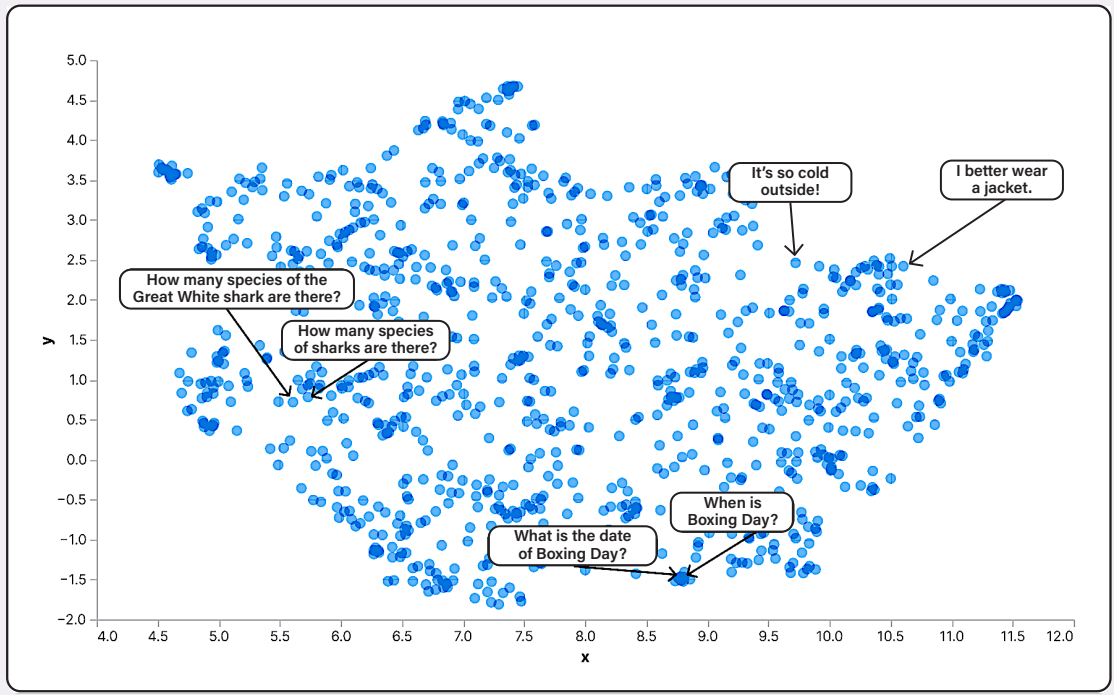

Technologists often try to address these shortcomings by training language models on more and more data. If a model is exposed to more data, the idea is that it will be familiar with more contexts, and outliers like the ironic cold-weather shorts blogpost will be outweighed by more representative data. This has led to ballooning in the size of large language models. BERT, a popular open-source model built by Google in 2018, was trained on 800 million words from free books and 2.5 billion words from English Wikipedia (Devlin et al., 2019). Two years later, OpenAI released its closed source GPT-3, which was trained on half a trillion mostly-English words crawled from the internet (Brown et al., 2020). Google's PaLM, released in 2022, trained on 780 billion words, mostly from English-language websites and social media conversations (Chowdhery et al., 2022). As models have grown in size, so have the computation costs of training them. While BERT costs a few thousand dollars in computing power to train from scratch and is often trained by academics to build new topic- or language-specific models (Izsak et al., 2021), GPT-3 and PaLM-sized models cost millions or tens of millions of dollars to train (Sharir et al., 2020). Future models will only be more expensive, leaving only the most well-off companies able to afford to build them (Bommasani et al., 2021).

Models are expensive to initially train, but once built, their representations are relatively cheap to use and be fine-tuned for different tasks. Thus, many technologists simply use pretrained large language models built by others (usually large companies, with the expertise and resources) instead of paying to create their own. The few big pretrained models that exist have thus become a sort of infrastructure, known as "foundation models" (Bommasani et al., 2021). This gives many technologists access to the state of the art capabilities, but it also creates a single point of failure for the sector as a whole: if a foundation model has a problem, it will persist across many applications. And these models are so large and complicated that even when they are open source, researchers cannot understand the underlying logic they use to come up with individual decisions.

Many of the largest and most advanced of these foundation models — such as OpenAI's GPT-4, Google's PaLM, and Meta's LLaMa — are trained primarily on English language data. In the next section, we explore one reason why that may be: the resourcedness gap.

## B. The Resourcedness Gap: Why the Largest Language Models are in English

English is the closest thing there is to a global lingua franca. It is the dominant language in science, popular culture, higher education, international politics, and global capitalism; it has the most total speakers and the third-most first-language speakers (Ethnologue, 2023b). It is the primary language spoken on the internet, accounting for 63.7% of websites, despite being spoken by only 16% of the world's population (Richter, n.d.). This dominance does not stem from any sort of inherent linguistic superiority: rather it is the colonial and neocolonial legacy of nearly three hundred years of the preeminent global superpower speaking English — first Great Britain, then the United States. The British government prioritized the English language through official language policies to facilitate trade and in an attempt to "modernize" its colonies, and as British, and later American trade became globally dominant, so too did English (Corradi, 2017; Phillipson, 1992). Prioritization of the English language came at the expense of other regional and indigenous languages and accelerated language endangerment and economic marginalization, which still impedes digital investment into these languages worldwide (Rowe, 2022; S. Zhang et al., 2022). American companies continue to perpetuate the dominance of the English language in a new more insidious form, by making online services available to global users without comparable investment into the languages they speak (Amrute et al., 2022; Kupfer & Muyumba, 2022).

As a result of these forces, English also dominates the field of natural language processing, and there is vastly more raw text data available in English than in any other language by orders of magnitude (Joshi et al., 2020). English has the most digitized books and patents, the largest Wikipedia, and the biggest internet presence. English is also by far the language paid the most attention by the global NLP research community. It is so hegemonic within the field that NLP papers about the English language typically do not even mention the language in the title or abstract (Bender, 2019). As Figure 2 shows, even among NLP papers that do mention a language in the abstract, English is mentioned over ten times as often as the next most mentioned language, German (ACL Rolling Review Dashboard, 2022).

This wealth of data and research makes it significantly easier to build large language models in English than in any other language. More raw text data, also known as *unlabeled data*, means more data for the model to be trained on; more research means that there are more datasets annotated with information, also known as *labeled data*, that can be used to test how well models complete different types of language tasks. This creates a virtuous cycle for English-language NLP — more labeled and unlabeled data leads to more research attention, which leads to increased demand for labeled and unlabeled data, and so on.

English is the prime example of a *high resource language*, a language for which a lot of high-quality data resources exist. Though it has the most data available of any language (English could be called an "extremely" high resource language), there are six other languages that could be considered high resource — the official UN languages list, minus Russian, plus Japanese (see Table 1). There are also a few dozen medium resource languages, such as Urdu, Italian, and Tagalog, with another one or two orders of magnitude less data, or about one hundredth or one-thousandth of available English data. The rest of the world's 6,000 plus languages can be considered low resource or extremely low resource, with only small amounts of written text available (Joshi et al., 2020).

Resourcedness can vary within languages as well. Languages such as Arabic and Spanish differ so much between dialects that many are mutually incomprehensible, even if they mostly use the same written form. Languages can also have different sociolects, varying across different social groups, identity groups, and contexts (e.g. formal versus informal). Regional dialects and sociolects can vary in degrees of difference from having different vocabulary and grammatical structures (e.g. Australian English or African American English versus Standard American English) to make extensive use of borrowed words from other languages (e.g. Nigerian English, Indian English), to fully hybrid bilingual dialects (e.g. Spanglish, Hinglish). However, the available digitized text of language often doesn't reflect the full spectrum of variation that exists within a language. (Bergman & Diab, 2022). Data scraped from the internet in particular over-indexes Standard English spoken by younger people in developed countries (Luccioni & Viviano, 2021). Other languages have just as much dialectical diversity as English and also likely over-index on certain dialects.

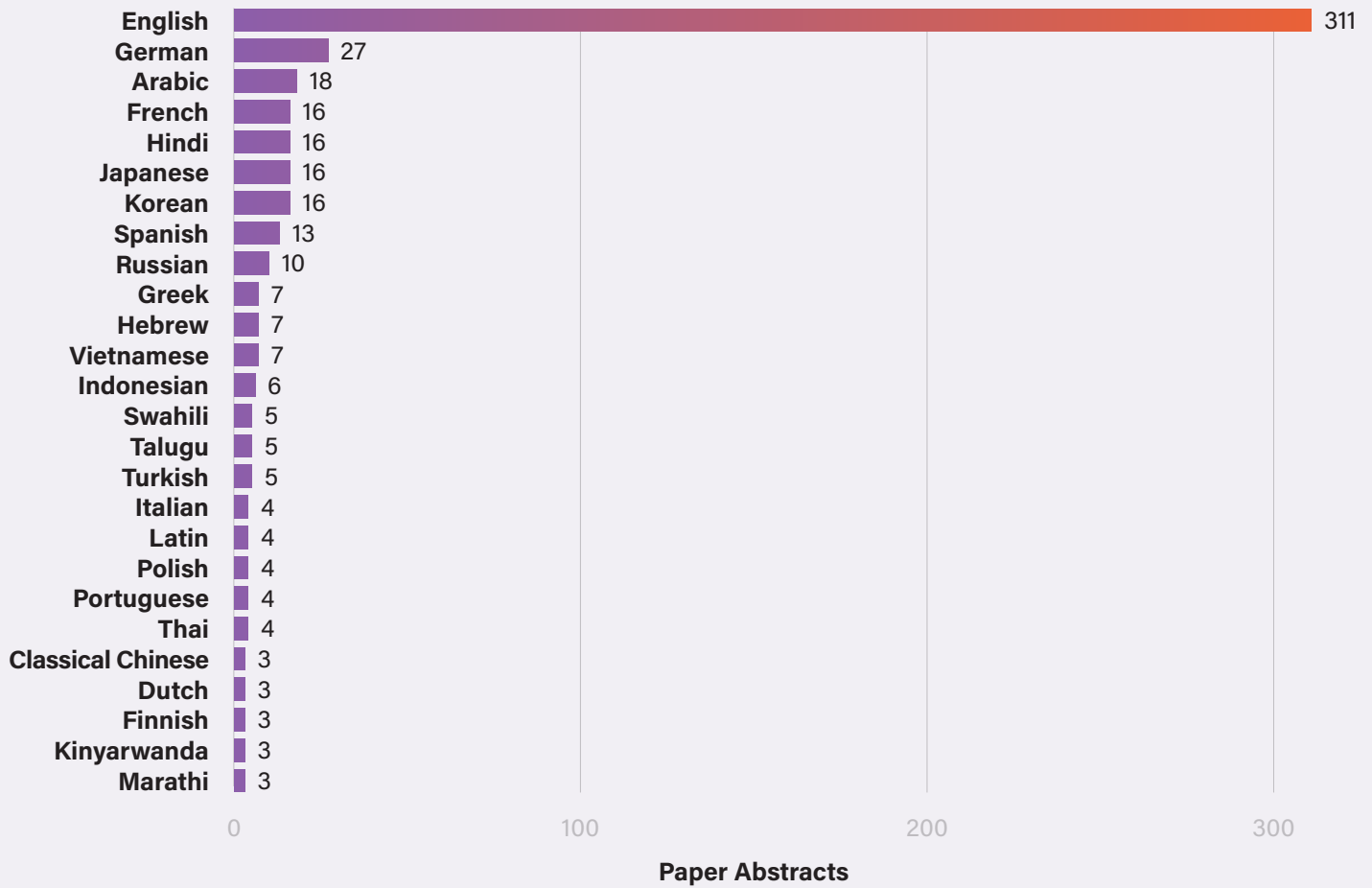

**Figure 2. Languages mentioned in paper abstracts.** Top most mentioned languages in abstracts of papers published by the Association for Computational Linguistics, May 2022-January 2023.

Source: (Santy et al., 2023)

Languages with less data available also often have lower quality data available, either because it is mislabeled or otherwise not representative of how people actually speak the language. This is particularly true with web-crawled data, a key data source for large language models (Khan & Hanna, 2023). Non-English language data scraped from the internet is more often machine translated, scanned from an image, or both, and each of those processes introduces opportunities for error (Dodge et al., 2021). Low- and medium-resource language data on the internet is more often pornographic, nonsensical, or non-linguistic content (Kreutzer et al., 2022). It is also often labeled as the incorrect language – around 95% of the time for many low resource languages – because automatic language identification works much more poorly with insufficient data, thus creating a circular problem (Caswell et al., 2020). Languages with the worst quality web data are disproportionately those written in non-Latin scripts (e.g. Urdu, Japanese, Arabic) and those spoken in the Global South (e.g. African languages, minority languages in the Middle East, non-Mandarin Chinese languages) (Kreutzer et al., 2022).

| Resourcedness | Languages | Number of Languages | Number of Speakers |
|---|---|---|---|
| Extremely High Resource | English | 1 | 1.1B |
| High Resource | Arabic, French, Japanese, German, Spanish, Mandarin | 6 | 2.7B |
| Medium Resource | Dutch, Vietnamese, Korean, Portuguese, Hindi, Slovak, Hebrew, Indonesian, Afrikaans, Bengali, etc. | Dozens | 2.7B |
| Low Resource | Haitian Creole, Tigrinya, Swahili, Bavarian, Cherokee, Zulu, Burmese, Telugu, Maltese, Amharic, etc. | Hundreds | 0.5B |
| Extremely Low Resource | Dahalo, Warlpiri, Popoloca, Wallisian, Bora, etc. | Thousands | 1.1B |

▲ **Table 1. Categories of language resourcedness.** Languages divided into different levels of resourcedness, according to labeled and unlabeled datasets available as of 2020.

Source: (Joshi et al., 2020)

Low resource languages also tend to have data that comes from a less diverse set of sources. The clean data that does exist often comes from places such as Wikipedia, the Bible, and parliamentary proceedings, particularly in large language models that depend on drawing parallels between low and high resource languages (see III.B and III.C) (Nekoto et al., 2020). None of these data sources is representative of a language as a whole. For example, there is a significant gender gap when it comes to who contributes to Wikipedia, with studies finding that the percentage of women who edit Wikipedia articles remains "dismally low" (Callahan & Herring, 2011; Vitulli, 2018), and it doesn't reflect a more casual style of speech. Some text on Wikipedia is also machine-translated — Cebuano, Swedish, and Waray for instance are some of the Wikipedia languages with the most articles, but most are translated by the same bot (Lokhov, 2021). The Bible is similarly its own unique domain, unrepresentative of language at large, but is overrepresented in the training data for non-English large language models. This can lead to errors in the tone and substance of language. For example, for a period of time, running a word repeated enough times through Google translate produced a religious-sounding text: the word "dog" pasted two dozen times and translated from Maori to English produced text about Jesus' return at the end of days (Christian, 2018).

The resourcedness of a language is often out of sync with the number of speakers or internet users that language has. Hindi, Bengali, and Indonesian are medium-resource languages yet each has hundreds of millions of speakers (Joshi et al., 2020). Guaraní, an Indigenous language spoken by most of the ~7 million-person population of Paraguay, hardly has any data resources at all (Góngora et al., 2021). Fula, a language spoken by tens of millions of West Africans, also has few data sets (Nguer et al., 2020). Despite over 600 million internet users across the African continent, nearly all African languages remain low-resourced.

Many scholars have worked to try to close this resourcedness gap between high and low resource languages. Individual NLP communities have formed around many languages in order to kickstart and perpetuate the virtuous cycle of research attention and benchmark development, including collectives such as IndoNLP for languages spoken in Indonesia and Masakhane for African languages (Cahyawijaya et al., 2022; Nekoto et al., 2020; Orife et al., 2020), and conferences such as the Association for Computational Linguistics' low resource language track, and AmericasNLP for indigenous languages (ACL, 2021; AmericasNLP, 2022; Masakhane, n.d.). Tech companies have also sought to expand the number of language models their models work in, in part by creating more data sets, including with projects like Facebook's No Language Left Behind project and Google's 1000 Languages Initiative (NLLB Team et al., 2022; Vincent, 2022). DARPA even funded the Low Resource Languages for Emergent Incidents (LORELEI) program in 2014 to improve translation about emergency incidents into low resource languages (Corvey, 2014). But the gaps between English, other high resource languages, and low resource languages remain large and are growing exponentially greater by the day, at least in terms of available, raw digitized data.

The response by the NLP community has not just been to collect more language data but also to employ technical tricks to help language models squeeze the most performance out of the little data they have. In the next section, we discuss the primary technical architecture developers use to do this: multilingual language models.

## C. Multilingual Language Models: Efforts to Bridge the Resourcedness Gap

In English, most large language models are *monolingual*, meaning that they train mostly on data from one language. Researchers have also built monolingual models in non-English languages: for instance, the architecture for Google's BERT model — one of the most popular and cheapest to train — has been utilized for French (CamemBERT), Italian (AlBERTo), Arabic (AraBERT), Dutch (BERTje), Basque (BERTeus), Maltese (BERTu), and Swahili (SwahBERT), to name a few (Agerri et al., 2020; Antoun et al., 2020; de Vries et al., 2019; G. Martin et al., 2022; L. Martin et al., 2020; Micallef et al., 2022; Polignano et al., 2019). However, in general, these monolingual models perform worse in their respective languages than the best English models do in English because they don't have as much data to train on.

This lack of data manifests in different ways depending on the specific task a model is fine-tuned to perform. Some language model capabilities — usually ones that depend on fact retrieval — improve linearly with size. For instance, the more data a language model is exposed to, the better it is at answering trivia questions or reformatting data (Srivastava et al., 2022). Other capabilities — usually ones with multiple steps or components — exhibit a "breakthrough" behavior, where once a model reaches a certain size, it improves sharply at the task. For instance, language models typically are unable to write code or add three digit numbers until they train on a certain amount of data, at which point their performance improves dramatically (Ganguli et al., 2022). Low and extremely low resource languages often do not have enough data to train a large language model at all, but medium and even high resource languages may not have the hundreds of millions, or billions of words of text data necessary to achieve the breakthroughs that English can (Y. Zhang et al., 2021).

Besides technical limitations, companies may not be interested in deploying a different monolingual model for every language their product is available in for business reasons as well. Maintaining and debugging one large language model for each language introduces costs that scale per language introduced, introducing complexity and additional overhead costs. Companies that seek to expand into new global markets will likely try to keep their costs fixed by reusing as much infrastructure as possible, including language models.

Therefore, instead of using monolingual models to do NLP tasks in non-English languages, researchers and developers most often use *multilingual language models*, such as Google's mBERT and Meta's XLM-R, which are trained on texts from many different languages at once. Like their monolingual counterparts, multilingual language models are trained on a fill-in-the-blank task. However, by training on text from several different languages, multilingual language models can, at least in theory, infer connections between languages, acting as a sort of bridge between high and low resource languages, allowing the former to bootstrap the latter.

For instance, imagine that an Indian climate change researcher wants to use a language model to collect all Hindi-language tweets about the weather. A monolingual language model trained on just Hindi text may not have enough data to have seen the words "thaand" ("cold" in Hindi) and "shaal" or ("shawl" in Hindi) appear near one another in text, so it may miss that tweets to the effect of "Main Agast mein shaal pahanta hoon" ("I put a shawl on in August") is a sentence about cold weather.[1] A multilingual model, trained on data from English, Hindi, and many other languages may have seen text where "thaand" appears near "cold," "shaal" appears near "shawl," and "cold" appears near "shawl," thereby allowing the model to infer that "thaand" and "shaal" are interrelated terms.

Multilingual language models are usually not trained on equal volumes of data from each language: mBERT for instance is trained on 15.5 GB of English text but as little as 10 MB of Yoruba text (Wu & Dredze, 2020). Even BLOOM, a popular multilingual model by BigScience with a particular focus on language representation, has 30% of its

---

1   Transliterated into Roman script for ease of reading for an English-language reader.

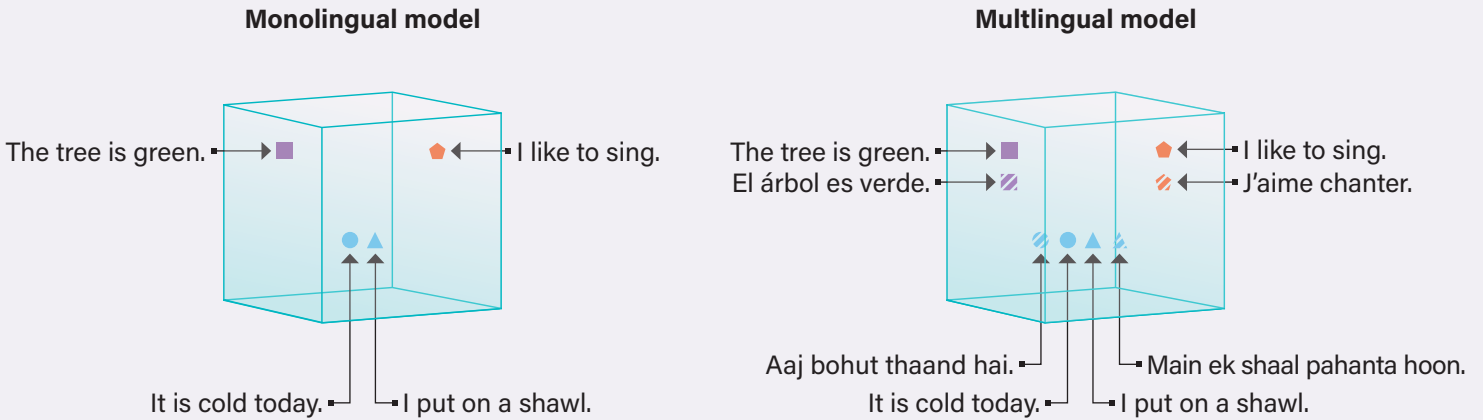

**Figure 3. Monolingual vs multilingual language model representation space.** A visualization of a monolingual and a multilingual language model's representation space, collapsed into three dimensions.

Source: (Schwenk, 2019)

training text in English (BigScience Workshop et al., 2023). In large part, this is because of the lack of available data in these languages, which come disproportionately from Wikipedia and religious texts, as discussed earlier (see Part I.C).

Just as a monolingual language model can be fine-tuned to work better on an individual task, a multilingual language model can be fine-tuned to work better in an individual language. Imagine for instance a developer who wants to use a multilingual language model to detect Indonesian election disinformation on social media. One way they could do it is by using an out-of-the-box multilingual model, such as BLOOM, and fine-tuning it by showing examples of false narratives circulated in Indonesian related to the local election. This likely would not work very well though, since BLOOM has only been exposed to a limited amount of data on Indonesian text — only 1.2% of its training data is in Indonesian (BigScience Workshop et al., 2023). Another better way to do it, if the developer has access to more Indonesian language data, would be first to fine-tune the model on additional Indonesian text (essentially, continuing to learn the fill-in-the-missing-word task, but this time just in Indonesian) and then further fine-tuning it on the task election disinfo detection using that dataset.

Model developers though do not always have enough text data to sufficiently fine-tune a multilingual model to work in a specific language. To make up for this, they often use imperfect machine-translated text. The two main methods of incorporating translated text are called *translate-train* or *translate-test* methods. With translate-train, a multilingual language model is fine-tuned on data that has been translated from (usually) English into a desired lower resource language (Conneau & Lample, 2019). With translate-test, a (usually) English monolingual language model is fine-tuned on data translated from the desired language into English, and all testing data gets translated into English as well (Artetxe, Labaka, et al., 2020).

Imagine, for example, a developer building a language model to detect terrorist content in the Basque language with a handful of examples of terrorist content in Basque but not enough Basque text data to properly fine-tune a language model. With the translate-train approach, a developer would take a large volume of English text data, machine translate it into Basque, use that data to fine-tune a pretrained multilingual language model, and then further fine-tune it to the task of terrorist content detection using the native Basque data. With translate-test, a developer would fine-tune a pretrained *English* language model on data translated from Basque to English, and then further fine-tune it by translating the terrorist content data they have into English. Subsequently, to analyze Basque text, it would first have to be translated into English before being evaluated by the model. Reliance on translated data raises many concerns, as discussed in Part II.C.1.

However, translated texts can help multilingual language models learn connections between languages. By feeding a model parallel texts — for instance, explicitly informing it that "baahar bohut thand hai" and "It's so cold outside" have the same meaning — it can better extrapolate other language parallels as well (e.g. NLLB Team et al., 2022; Reid & Artetxe, 2022). Multilingual language models can learn connections between languages without explicit labeling, instead inferring relationships between languages on its own through borrowed words, numbers, and URLs (Pires et al., 2019).

In general, NLP researchers understand little about why it is that multilingual language models can be effectively fine-tuned to work in languages that they have relatively little data for (Conneau, Khandelwal, et al., 2020; Pires et al., 2019; Wu & Dredze, 2019). Some argue that it is because multilingual language models have inferred language-agnostic concepts and universal rules that can be applied to any language (Artetxe, Ruder, et al., 2020; Chi et al., 2020; Conneau, Wu, et al., 2020; Tsvetkov et al., 2016). Others say that multilingual language models are just effective imitators (Bender et al., 2021; Lauscher et al., 2020). The debate is impossible to fully resolve because of the overall complexity and opacity of large language models, but so far evidence suggests that at best, the linguistic universals they learn are limited to narrow semantic and syntactic domains (Libovický et al., 2019; Wu & Dredze, 2019), such as learning plural/singular verb agreement across multiple languages (de Varda & Marelli, 2023). But even if a model can infer syntactic or semantic commonalities between languages, such inferences will not necessarily help it manage more complex, context-dependent tasks (Choi et al., 2021). For instance, in some languages, multilingual language models do no better than random guessing at detecting hate speech (Lin et al., 2022). As will be discussed in the next section, these are hardly the only limits of multilingual language models.

# II. Limitations of Language Models in English and Non-English Contexts

The press, technology companies, and social media are abuzz about the potential of large language models. In this section, however, we discuss the shortcomings of these models, particularly as they operate in non-English language contexts. In the first section, we discuss general concerns with building and deploying large language models. These concerns apply both to the English and non-English contexts. In the second section, we look at the problems more specifically raised by multilingual language models.

## A. Concerns with Building and Deploying Large Language Models

### 1. LARGE LANGUAGE MODELS ARE BOUND BY LANGUAGE THEY HAVE SEEN BEFORE AND STRUGGLE TO PERFORM IN NEW CONTEXTS.

A large language model does not understand language; instead, it makes probabilistic inferences about text based on the distribution of language within the data it is trained on. Bender and Koller argue that this means language models are limited to contexts they have encountered before and struggle greatly in those they have not (2020). NLP researchers have already proven this is the case in generative AI by demonstrating several unintuitive outcomes: for instance, language models are better able to perform mathematical operations with numbers that appear frequently in written language (e.g., multiplying numbers by 24), than numbers that appear infrequently (e.g. multiplying numbers by 23) (Razeghi et al., 2022). Large language models may exhibit similar limitations in content analysis as well. For instance, if a large language model were used to analyze a candidate's resume, it may struggle to account for lesser-known companies or newer skill sets without up-to-date, domain-specific data to fine-tune on. These tasks are reliant on in-context knowledge and without domain-specific training, i.e. training an off-the-shelf large language model with text relevant to the task at hand, these models are likely to perform poorly and their purported domain-agnostic abilities should garner skepticism (Duarte et al., 2017).

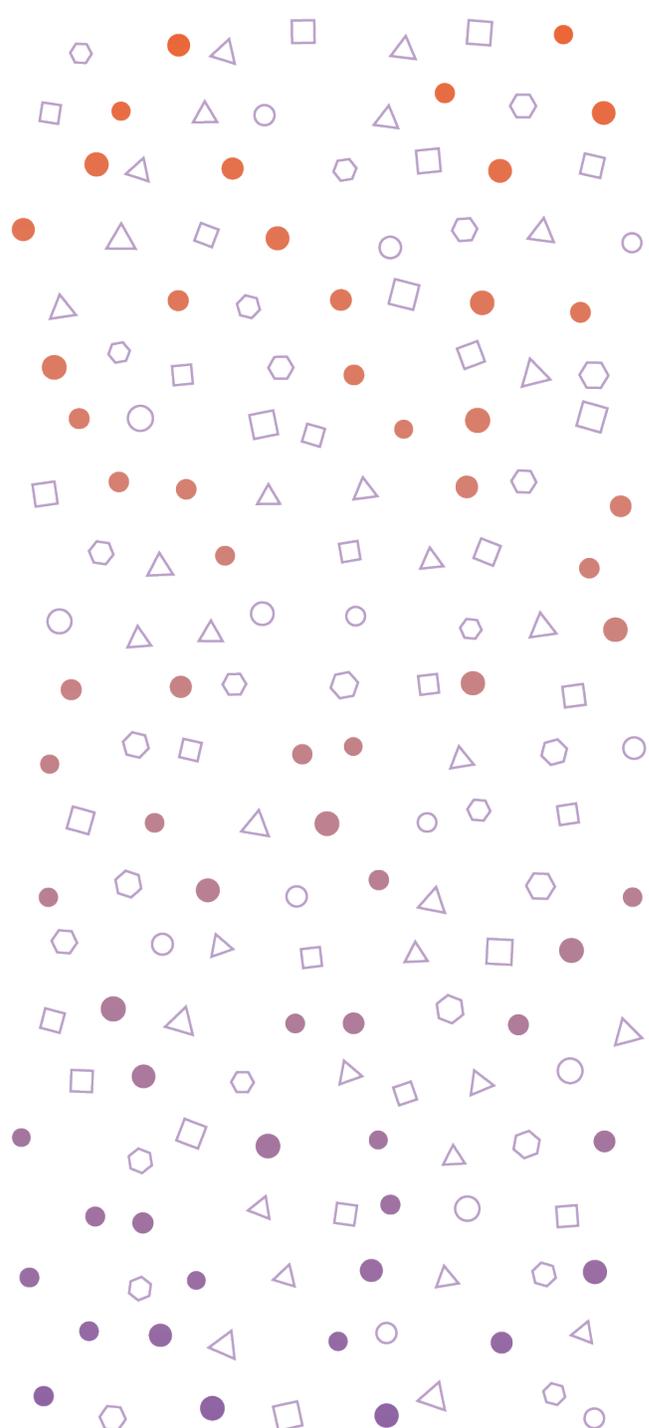

## 2. LARGE LANGUAGE MODELS REPRODUCE THE BIASES, VALUES, AND HARMS OF THE DATA THEY TRAIN ON.

Large language models are built using vast quantities of text scraped from the internet and exhibit all the biases and limitations of their data source (Okerlund et al., 2022). Some commonly used datasets, such as Common Crawl, include large volumes of hate speech and sexually explicit content (Luccioni & Viviano, 2021). Other problems are more nefarious. For example, researchers found that when GPT-3 generated completions for the prompt "Two Muslims walked into a____," 66% of completions included violent language, three times more than for other religious groups (Abid et al., 2021). Others have found similar entrenched biases against people with disabilities, for example inferring negative sentiment from sentences that include disability-related terms (Hutchinson et al., 2020).

Though technologists often try to pull out explicitly harmful data from training sets, models can still reify harms, such as referring to "women doctors" or calling undocumented immigrants "illegals" (Bender et al., 2021). Removing these instances of harmful data from training datasets, which are disproportionately outsourced to underpaid staff around the world, also imposes labor and psychological burdens (Williams et al., 2022).

Even if datasets are rid of specific examples of harmful text, they will nonetheless contain values and assumptions that are encoded into the language we speak and the dominant perspectives that exist in many pieces of written text, particularly government documents or state-run media pieces that may make up the bulk of text available for low resource languages (Bender et al., 2021). Many machine learning researchers fail to consider these problems in their work — one study found that 98% of machine learning papers mention no negative potential of the technologies they are describing (Birhane et al., 2022). Yet the risks are very real: as Birhane & Prabhu put it, "Feeding AI systems on the world's beauty, ugliness, and cruelty, but expecting it to reflect only the beauty is a fantasy" (2021). When these problems exist in any particularly popular foundation model, they proliferate across many different applications built on top of that model.

## 3. THE DATA LARGE LANGUAGE MODELS TRAIN ON RAISE COPYRIGHT AND PRIVACY CONCERNS.

Legal experts also raise concerns about copyright and ownership of text that make up the vast quantities of data that train and distinguish large models (Ebers et al., 2022; Okerlund et al., 2022). Getty Images has sued the creators of Stable Diffusion, an AI tool that creates images based on written prompts, claiming that the toolscraped Getty's databases of proprietary images and photos without permission (Vincent, 2023a). Legal questions about ownership of text and whether scraping proprietary text is lawful (e.g., because it constitutes fair use) or not remain unanswered (Kublik, n.d.).

Some datasets that large language models train on are likely to capture examples of language from sites such as social media, raising personal data privacy concerns. There is a high possibility that in gathering exchanges from social media networks, training datasets inadvertently contain private and even sensitive information, which increases the risk of models leaking details like names, phone numbers, or addresses from the data on which they're trained (Carlini et al., 2021, 2023).

### 4. TRAINING LARGE LANGUAGE MODELS COULD HAVE A SIGNIFICANT ENVIRONMENTAL IMPACT.

Finally, there are increasing concerns about the environmental cost of producing large language models. Scholars and advocates have raised concerns about the environmental impact of training these models, particularly the largest ones with billions of parameters, due to their intense computation requirements (Crawford, 2021; Okerlund et al., 2022). There is preliminary research attempting to quantify the energy impacts of computation at this scale (Kaack et al., 2022), but some early estimates suggest that training a single BERT model, one that serves as the foundation for some multilingual language models, requires as much energy as a trans-American flight (Strubell et al., 2019). Large language models, like GPT-3, require thousands of times more (Heikkilä, 2022). Png writes that these costs may be concentrated in poorer countries, where server farms and raw materials required to build necessary infrastructure are often located (2022).

# B. Limitations of Multilingual Language Models

## 1. MULTILINGUAL LANGUAGE MODELS OFTEN RELY ON MACHINE-TRANSLATED TEXT THAT CAN CONTAIN ERRORS OR TERMS NATIVE LANGUAGE SPEAKERS DON'T ACTUALLY USE.

Incorporating machine-translated data into the training and fine-tuning of multilingual language models creates various opportunities for the model to malfunction. Multilingual language models that depend on translation may struggle to build accurate representations of words or concepts which have different connotations in different languages. For instance, in English, "dove" is a term associated with peace, but its equivalent in Basque, "uso," is an emasculating insult. A translation-based cross-lingual model that does not train on the word "uso" used in its native context could potentially fail to see it used in a call for violence since the English mapping is so closely associated with "peace."

Another issue is what NLP practitioners call the "translationese" problem (Yu et al., 2022) — that is, machine-translated language materially differs from how human native speakers naturally use language (Bizzoni et al., 2020; Teich, 2003). In generative AI, translationese can result in mono- or multilingual language models simplifying or overcomplicating sentences, producing repeated words, using too common or too uncommon words, borrowing too much or too little from the original language, and other patterns of speech native speakers would not use (Volansky et al., 2015). These mistakes are not consistent between languages or systems, so it would be difficult for models to be able to systematically root them out, though some argue that it is possible (Yu et al., 2022).

The problems of machine translation spread beyond models that intentionally train on it. The web is filled with machine-translated text, and models that train on web-scraped data will inadvertently encounter a lot of it, particularly in low resource languages (Kreutzer et al., 2022). For instance, a lot of the Catalan data that exists on the web, particularly on websites using the .cat top-level domain, is translated using Google Translate, even on official government websites (Pym et al., 2022). Even benchmarks to test how well multilingual language models work in high and low resource languages are often translated from another language, leaving researchers with less of a sense of how well these models work on language as spoken by native speakers. For instance, OpenAI tested GPT-4's capabilities in 26 languages, but using only benchmarks translated from English (OpenAI, 2023).

## 2. MULTILINGUAL LANGUAGE MODELS FAIL TO ACCOUNT FOR THE CONTEXTS OF LOCAL LANGUAGE SPEAKERS.

As discussed earlier, large language models only work well in contexts similar to contexts of the data they are trained on. A language model trained on legal texts, for instance, will perform much better on law-related tasks than medical tasks or interpreting the Quran (Koehn & Knowles, 2017). This poses a problem for multilingual language models, which, particularly in low resource languages, are trained on text that is translated from other language contexts or comes from a few distinctive contexts, such as Wikipedia and the Bible. Multilingual language models that are not trained on large volumes of text from native speakers of a given language will more often fail at tasks that require knowledge of an individual speaker's local context, such as hate speech detection and resume scanning (Lin et al., 2022).

Imagine, for example, a multilingual language model fine-tuned to detect anti-Muslim content in Assamese, a low-resource language with fifteen million speakers, predominantly in northeast India (Ethnologue, 2023a). Assamese and Bengali are both medium resource languages, so a multilingual model may draw connections between the two. However, anti-Muslim hate speech is very closely tied to historical events and the specific political conditions of Assam. For instance, the term "Bangladeshi Muslim,"

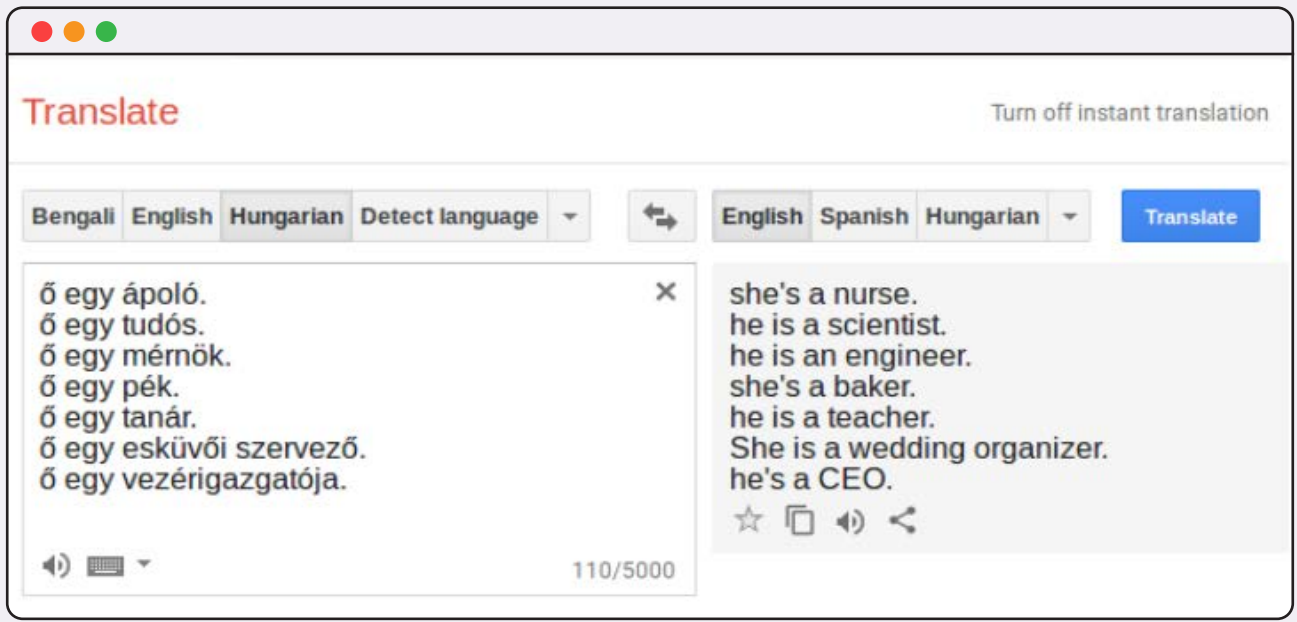

**Figure 4. Google Translate from Hungarian to English.** A screenshot of Google Translate, circa 2020, showing how the multilingual language models project gender onto genderless languages.

Source: (Prates et al., 2020)

neutral in many other languages and contexts, is a hate speech dog whistle in Assamese because it casts Assamese Muslims as foreigners (a concept that is itself closely tied to the Indian government's repatriation efforts) (Avaaz, 2019). A multilingual model neither trained on extensive native Assamese text nor explicitly trained by a language expert would likely not be able to capture this hyperlocal distinction.

Multilingual language models work by transferring between language contexts, but that transfer often means simply that the context of higher resource languages overwrites lower resource ones. Spanish, for instance, tends to use more adjectives and analogies describing extreme situations than English, so a sentiment detection algorithm that transfers linguistic properties over from English may mischaracterize Spanish text as having a stronger emotional valence than it would to a native speaker (Stadthagen-Gonzalez et al., 2017). This structure transfer can also bring the biases of a source language into a target language (Savoldi et al., 2021). For instance, if a language without gender pronouns, such as Hungarian or Yoruba, is mapped onto a language with gendered third-person pronouns, such as English or French, the language model could force gender associations and biases of the gendered language onto the non-gendered one, as often occurs in translation (Prates et al., 2020) (see Figure 4).

### 3. MULTILINGUAL LANGUAGE MODELS DO NOT AND CANNOT WORK EQUALLY WELL IN ALL LANGUAGES.

Multilingual language models not only do not work equally well in all languages but they cannot, since the more languages a multilingual model is trained on, the less it can capture unique traits of any specific languages. This problem is called the *curse of multilinguality* (Lauscher et al., 2020). Large language model developers are thus forced to trade off performance between disparate languages; making a model work better in Hindi for example, may come at a cost to its performance in English. In practice, when technology companies must choose which languages to deprioritize within their multilingual language models, they may be incentivized to have them be languages where speakers tend to be less wealthy, have less political power, or live outside of the company's priority markets, thus exacerbating the resourcedness gap they are designed to address.

In general, semantic and syntactic similarity to a high resource language protects from the curse of multilinguality (Eronen et al., 2023). For instance, Muller et al. tested mBERT on languages it had not explicitly trained on before and found that it worked better in Swiss German (related to German, a high resource language), than it did in Estonian (a Uralic language, like medium resource languages Hungarian and Finnish), than it does Uyghur (a Turkic language, distant from any high or medium resource language, with four alphabets) (2021). In general, multilingual language models struggle with languages written in non-Latin scripts (Pires et al., 2019; Ruder et al., 2021), language isolates (languages etymologically distinct from all other languages, such as Basque), and families of languages less connected to those of high resource languages. This threatens to create a poor-get-poorer dynamic for languages that are only similar to other low resource languages, as is the case with many widely spoken African languages including Swahili, Amharic, and Kabyle (Joshi et al., 2020). This dynamic further strengthens the post-colonial structural inequality discussed throughout this report.

Multilingual language models are also forced to trade off between languages in the vocabulary they use. Large language models train on the problem of predicting the next word in a sentence. If a model is trying to guess the word to fill in "Today I feel ___," it will have a harder time doing so if it has to choose between ten million possible words from any language instead of just a few hundred thousand English words. The total number of words a language model has to choose from is called its vocabulary size. The larger a model's vocabulary size, the more different possible words it can generate and recognize, but also the more computational resources it takes to train. Multilingual language models use all kinds of shortcuts to get their vocabulary size down. For instance, they will often transliterate languages into Latin scripts or train the model to guess the next subword (e.g. breaking "tasks" into "ta" and "##sks") or letter instead of the full word, thus collapsing the barrier between languages (Tay et al., 2022; C. Wang et al., 2020). These shortcuts cut down on costs, but they also reduce a model's ability to capture semantic relationships between words, thus degrading its performance overall.

Vocabulary is often decided by how frequently different words, subwords, and letters appear in a model's training text, and since multilingual language models are trained mostly on English data, their vocabularies will skew towards English as well. A multilingual model may have a relatively obscure word like "riposte" in its English vocabulary, but be may missing common words in other high resource languages (e.g., "escritorio" in Spanish), common subwords in medium resource languages, (e.g., "tzv" in Hebrew), and entire letters in low resource languages (e.g., a character that appears in Tigrinya but not other Ge'ez-based scripts). This inferior representation makes models perform worse in a variety of tasks, and makes content analysis systems far easier to trick by doing things like changing white space, using typos, or in the case of toxic content detection, adding common, positive words like "love" (Gröndahl et al., 2018; Lees et al., 2022).

## 4. WHEN MULTILINGUAL LANGUAGE MODELS FAIL, THEIR PROBLEMS ARE HARD TO IDENTIFY, DIAGNOSE, AND FIX.

NLP practitioners depend on benchmarks to determine both how well a language model performs at specific tasks and how close it is in general to achieving "natural language understanding" (Bender & Koller, 2020). This latter type of benchmarking is very difficult in all languages, since it is hard to generalize about a language model's capabilities from only a handful of disparate tests (Raji et al., 2021). However, the challenges of both types of benchmarks are exacerbated in the multilingual context. The disparities in NLP research attention and labeled data between languages mean that there are far more benchmarks and tasks that can be used to test models in English than in other languages, particularly low resource ones. Models developed to operate in non-English contexts are still usually tested with benchmarks translated from English which, as discussed earlier, is often markedly different from the target language.

The alternative to translation is hiring people local to the contexts a model is being applied to and paying them to create data sets and develop benchmarks. This works particularly well for models built to do a specific task in a specific language (Nguyen, 2020; Tattle, n.d.), but is very expensive and resource intensive to scale up for models meant to work in many languages and contexts. It also raises challenging questions for detecting bias in language models (Talat et al., 2022) and performing inherently political tasks, such as content moderation. For instance, a social media company trying to create a dataset of inflammatory content posted in Bosnia and Herzegovina needs people who are experts in multiple ethnic conflicts and languages (Bosnian, Serbian, Montenegrin, and Macedonian) but also unbiased in those conflicts, all in a country that lacks media pluralism or a strong civil society sector (Article 19, 2022). Scaling this to every geopolitical problem discussed in all languages on a given online service is a daunting, if not impossible, task.

When problems with multilingual language models can be found, it is often difficult to determine why they are occurring. Large language models are already opaque, even to those who develop them — neural networks, the core technology underlying large language models, are known for being particularly obtuse and for representing language in a way that doesn't map cleanly onto human-understandable concepts (Nicholas, 2020). However, multilingual language models are particularly opaque because they make unintuitive, hard-to-trace connections between languages. Take for instance, this case from an NLP paper: the Google researchers behind the Perspective API, a model for detecting "toxic" content, found that their model flagged tweets that used the Italian word "sfiga" (which roughly translates to "bad luck") as hate speech because two of the three examples included in the training dataset that contained the subword "sfiga" were labeled as hate speech ("sfigati" is an insult meaning "loser") (Lees et al., 2020). If this were a multilingual model that had mapped Italian learnings onto Turkish analysis, perhaps sentences with the equivalent Turkish word for "unlucky" ("şanssız") would also be flagged as hate speech. Even if researchers had access to all the data used to train that multilingual model, it would be extremely difficult to locate and fix this bug without knowing Italian or understanding how the model had mapped these relationships.

# III. Recommendations

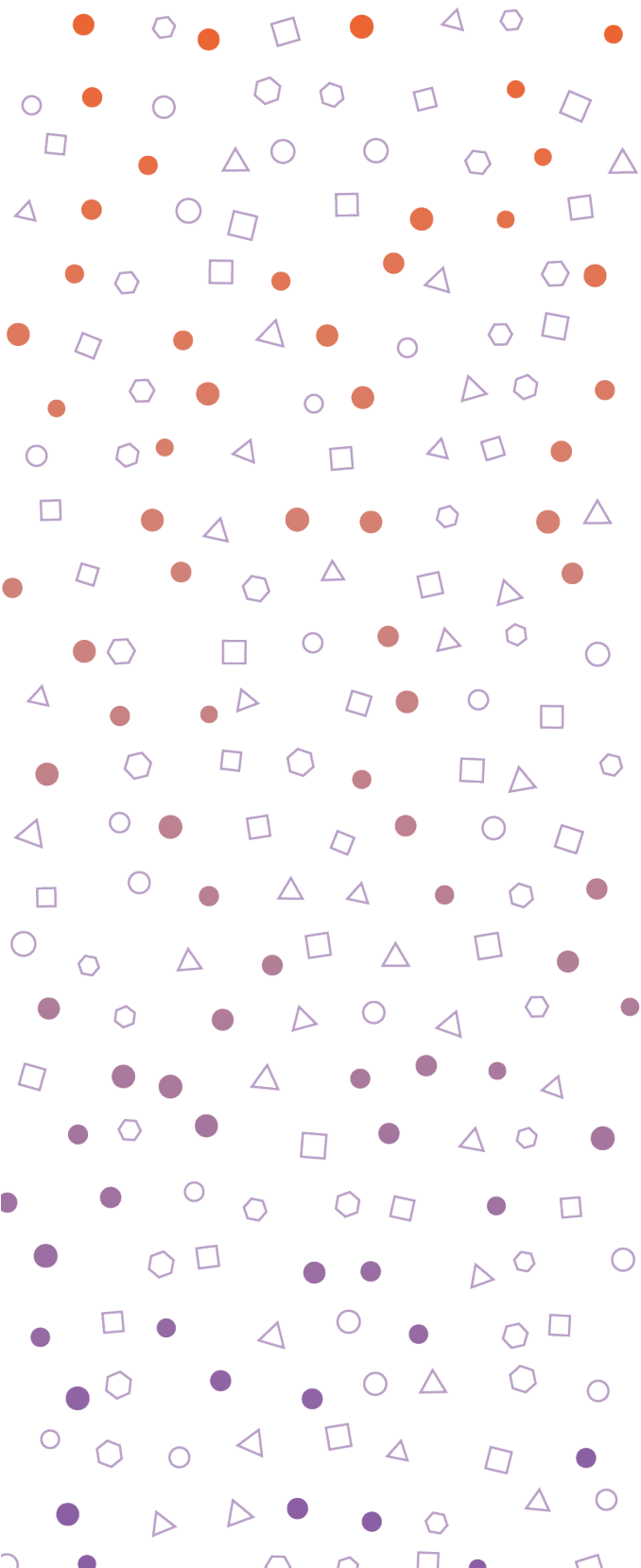

Efforts to improve language models' performance in various languages and contexts are exciting, as they may boost connectivity and information exchange for billions of users around the world. However, language models are limited in their capabilities, and employing them too widely, without safeguards, or for the wrong kinds of tasks has the potential to raise civil liberties concerns and erect new barriers (Maundu, 2023). Unthinking deployment of large language models may impede peoples' ability to access information, employment, and public benefits, with disparate impacts for individuals in the Global South where many of the low resource languages are spoken. We should be cautious about the rapid adoption of these technologies, especially as building blocks for other types of automation in high-stakes arenas like content moderation, employment software, and resource allocation.

In this section, we offer recommendations for companies, researchers, and governments to take into consideration as they build, study, and regulate large language models, particularly in non-English language contexts.

## A. Companies

### TECHNOLOGY COMPANIES SHOULD DISCLOSE WHEN, HOW, AND IN WHAT LANGUAGES THEY USE LARGE LANGUAGE MODELS.

To better understand the problems and challenges with deploying large language models in different languages, researchers and the public need to know where to look. Companies that incorporate language models into their technical systems should always disclose how they are using them, which languages they are using them in, and what languages they have been trained on. Currently, the approach of many companies to AI transparency consists of trumpeting the capabilities of their AI systems in blog posts and press releases, and, for a few larger firms, releasing research versions of their language models that still differ from the ones they use in production. Despite publishing on AI and pushing the field forward, technology companies tend to hold information about their production AI systems, even basic information about what languages they are used in, close to the chest.

Academics and civil society have written extensively about how technology companies, particularly online service providers, could offer better transparency and accountability for their AI systems, including language models. The Santa Clara Principles, a set of principles developed and revised by global civil society groups, provides examples of the types of disclosures companies can make about their content moderation policies and processes (2021). Groups like BigScience also pave the way, exemplifying the type of documentation other model-developers can publish about their content analysis systems, including model cards, transparency reports, and other avenues to disclose more information about the linguistic makeup of a model's training data (e.g. what languages it has trained on, how much data from each language, where those datasets come from). Better transparency creates opportunities for external actors to more immediately identify potential risks and impacts on users and for technology companies to mitigate the potential dangers of deploying large language models in English and non-English contexts.

**WHEN DEPLOYED, LARGE LANGUAGE MODELS SHOULD BE ACCOMPANIED BY ADEQUATE REMEDIAL CHANNELS AND MECHANISMS THAT ENSURE INDIVIDUALS CAN APPEAL OUTCOMES AND DECISIONS MADE BY THESE SYSTEMS.**

Because of the complexities of human speech and the error-prone nature of automated tools, decision-making systems built on top of large language models should be used within narrow remits and with adequate remedial channels for users encountering them. Those remedial channels and processes should have human reviewers with the same language proficiencies that their systems are deployed in. Language- and context-specific remedial channels are particularly important for allowing users to appeal decisions made by online services, especially when those decisions either restrict their expression or access to information or fundamentally determine their access to economic or social rights like the right to housing, education, and social security (United Nations Human Rights Office of the High Commissioner, n.d.).

Technology companies can also offer accountability at a system level, not just the level of individual decisions. One way to do this is to conduct and publish human rights impact assessments at the different phases of the language model's life cycle — development, testing, deployment, and evaluation (Prabhakaran et al., 2022). Publishing human rights impact assessments will also aid in other actors' decisions when procuring these systems to conduct tasks in different domains and contexts. In particular, these human rights impact assessments should consider the disparate risks to different language speakers in advance of a model being deployed in those languages. Online service providers can provide transparency by disclosing the systems and languages they use large language models in.

**COMPANIES SHOULD INVEST IN IMPROVING LANGUAGE MODEL PERFORMANCE IN INDIVIDUAL LANGUAGES BY BRINGING IN LANGUAGE AND CONTEXT EXPERTS.**

Recently, an arms race has begun between Google and Meta to see who can include more languages in their multilingual language model. Meta's "No Language Left Behind" initiative trained a model on over 200 languages (NLLB Team et al., 2022); months later, Google one upped Meta with its "1,000 Languages Initiative" (Vincent, 2022). This race puts a premium on the number of languages the model trains on, rather than how well it works in each language. In particular, it is unclear how these models will handle the "curse of multilinguality," where, as explained in II.B.3, the more languages a model trains on, the less it can capture the idiosyncrasies of each language. It is also unclear how these companies define a model "working" in any of these languages.

Companies building large language models should not just focus on the number of languages their model is trained on but the quality of its performance in each language. In part, that means better benchmarks, but benchmarks can only go so far. To evaluate the full range of potential applications and pitfalls that could come with applying a language model in a specific language context, it is necessary to involve language experts, civil society, local experts, heritage and language preservation advocates, linguists, and human rights experts. These actors are crucial to ensuring that labeled training datasets adequately capture the nuances and variations of a given language. Many organizations are already doing this type of work. Uli is an example of this, where two India-based nonprofit organizations — Tattle and Centre for Internet & Society — convened a range of gender, gender-based violence, communal violence, and other language experts to annotate training datasets in Indian English, Tamil, and Hindi to build a tool capable of parsing sentiment and toxicity on Twitter. Other researchers have also pointed to using annotators to label training datasets as a way to equip models with the ability to parse variations in the speech of a certain language (Bergman & Diab, 2022; Nkemelu et al., 2022).

# B. Researchers and Funders

**RESEARCH FUNDERS SHOULD INVEST IN SPECIFIC NLP LANGUAGE COMMUNITIES TO KICKSTART THE VIRTUOUS CYCLE OF DEVELOPMENT.**

Developing NLP capabilities in any language is a cyclical process, and for high resource languages — particularly English — that cycle is virtuous. When a language has lots of clean, human-annotated datasets, researchers and developers are better equipped to build models and benchmarks to test models in that language. More models and benchmarks lead to more publications, conferences, and real-world use cases. And finally, increased demand for research and software in a language drives demand for more datasets. For low resource languages, however, the virtuous cycle is hard to kickstart. Without tools, annotators, and financial investment earmarked for different language communities, NLP researchers cannot create the datasets needed to build models or benchmarks, and even if they could, they face difficulties publishing or getting attention for their work in popular journals and conferences. The most prestigious NLP publications focus disproportionately on English; languages without their own self-sustaining NLP communities end up to a handful of specialized outlets.

Investments into non-English NLP should particularly focus on creating self-sustaining scholarly NLP communities, and doing this requires investing in all levels at once. The groups that are best set up to properly allocate these investments are the language- and geography-specific NLP research communities that have cropped up over the years, such as such as Masakhane, AmericasNLP, ARBML, and others who can convene practitioners around common goals to advance the field (Alyafeai & Al-Shaibani, 2020; AmericasNLP, 2022; Orife et al., 2020). These communities know what kind of data sets should be built, which community actors are needed to properly vet them, and what kind of competitions and conferences should be run to keep the virtuous cycles going. One model for how this can work is exemplified by EVALITA, an event hosted by the Italian Association for Computational Linguistics. In it, researchers first submit data sets for new language tasks, such as identifying misogyny or dating documents. Then, researchers compete to train models to perform those tasks the best. Finally, those results get published, thus generating interest and attention toward Italian NLP and ensuring researchers continue to build tools for the language (Basile et al., 2020).

Private companies can contribute not only by financially supporting these efforts but by sharing more of the non-English datasets they use to train their large language models, both for transparency and to support research. Large tech companies have already shared the code for training many of their multilingual language models — Meta's XLM-R and Google's mBERT are the subjects of most multilingual model research in publication — and disclosed the data they train them on — CommonCrawl, and Wikipedia and BooksCorpus, respectively. However, the models that Google, Meta, OpenAI, and other large companies use in their products train on other, proprietary, language data. Companies should share more of their training data, both for public accountability and to bolster research.

Large language models have by and large been built by private companies, but private incentives may be at odds with developing these models in safe and equitable ways. Government investment into non-English large language model research could lead to improvements in areas private companies may be underinvesting in (Mazzucato, 2014). DARPA's late 2010's LORELEI project, aimed at spurring research into low resource languages to improve translation for humanitarian efforts, is a good first step, but further government incentives could help assure that NLP researchers invest in a broad range of approaches and languages, rather than focus disproportionately on English. BigScience's BLOOM is a good example of how large language models can be developed in the open and with public support. The French government is one of many funders which has allowed BLOOM to remain open to inquiry by other NLP practitioners. The multilingual language model was trained using ROOTs, a 1.6TB multilingual dataset that is clearly documented and available for NLP practitioners to analyze (Laurençon et al., 2022).

## RESEARCHERS SHOULD FOCUS ON MEASURING AND ADDRESSING THE IMPACTS OF LARGE LANGUAGE MODELS.

Technologists understand little about the internal logic of how large language models operate and therefore have a difficult time predicting when they make mistakes, what the effects of these mistakes will be, and how to fix them. Multilinguality only exacerbates this problem. Better tools are needed to interrogate large language models, particularly multilingual language models, about why they make the decisions and mistakes they do, and how to fix them.

In particular, the increased use of multilingual language models has the potential to help and harm language communities. Enabling greater digital participation amongst a language community raises something that researchers call the "Janus-face nature of digital participation" (NLLB Team et al., 2022): it allows more to participate and benefit from the digital economy, however, it may also expose more people to the harms present online, often without their consultation and consent (Hao, 2022; Toyama, 2015). More research on the effects and externalities of the increased use of language models and specifically multilingual language models must grapple with the impacts these tools have on different linguistic communities, linguistic preservation and diversity efforts, and access to opportunity for all.

Different actors have different roles to play here. Civil society has a role in documenting the impacts of these models and imagining what these "better" models should look like. There are many open questions around the types of problems that need automated solutions, what more representative datasets might look like, how to manage the tradeoffs between languages, how large language models affect linguistic preservation efforts, and what the rights implications are of using large language models, among other things. Academics and corporate researchers have a role in better defining the contexts and tasks these models hope to address, and developing quantitative and qualitative methods to evaluate these desired normative values. And companies that deploy language models can provide researchers more transparency into how their models work, what data they are trained on, and in what situations they use them so researchers can better tailor their research to reflect what is happening in real-world systems.

# C. Governments

**GOVERNMENTS SHOULD CAUTION AGAINST USING AUTOMATED DECISION-MAKING SYSTEMS THAT RELY ON LARGE LANGUAGE MODELS TO MAKE HIGH-STAKES DECISIONS.**

Many governments have deployed or are considering deploying systems that use natural language processing technology as part of AI systems to make high-impact decisions, such as determining immigration status or selecting judicial cases to try (Patel et al., 2020; Rionda & Mejia, 2021). Vendors who build these systems may soon follow the larger industry trend of incorporating large language models since they are relatively cheap to build and easy to adapt as requirements change. However, as discussed throughout this paper, large language models are a relatively novel technology that has technical limitations. These tools pose serious civil liberty concerns that are magnified in non-English contexts and when used to make decisions that may affect a person's livelihood. For instance, if a large language model is used as the basis of an algorithm to evaluate affordable housing applications and the text that large language model was trained on exhibits anti-Muslim bias, the resulting affordable housing algorithm may disproportionately deny Muslims' applications. Relying on large language models to make high-stakes decisions can have outsized, negative impacts on individuals' lives, impeding safety and access to economic opportunities.

Governments should therefore never rely solely on automated systems that incorporate large language models to make high-risk decision-making areas, such as pretrial risk assessment, allocation of social services, and immigration status. Policymakers should consider the impact on rights and access to services when procuring new tools and vendors to build these systems and conduct and disclose any assessments conducted on these systems. They should also be cautious when adopting these systems for information sharing services, such as chatbots about social services or that provide healthcare information, and test them extensively in every language in which they are deployed, and never use them to entirely replace human intermediaries.

**GOVERNMENTS SHOULD NOT MANDATE OR INADVERTENTLY REQUIRE BY LAW THE USE OF AUTOMATED CONTENT ANALYSIS SYSTEMS TO DETECT OR REMOVE CONTENT IN ANY LANGUAGE.**

Governments around the world are increasingly pressuring online service providers to limit content they find to be inaccurate or harmful, such as misinformation related to health care, or preemptively monitor online speech which may incite violence. Given the scale of content available on social media and other services, this has driven an interest amongst governments to mandate that online service providers use automated content analysis systems to detect or remove content they deem as "illegal" or harmful to their constituents.

This is ill-advised. Mandating the use of automated content moderation technologies or requiring companies to take down content in a limited time period (effectively requiring the use of automated technologies) opens the door for the overbroad removal of speech. Large language models, especially in non-English language contexts, are not a magical technology that can perfectly distinguish between "good" and "bad" speech. At best, they are an imprecise technology that fails to understand the context of speech — for instance, when an individual uses a slur versus when a journalist documents the use of a slur by that individual. At worst, they are tools that can be appropriated by governments to squash dissent and freedom of expression. Efforts to persuade tech companies to improve their automated systems, clarify their policies, introduce more accountability, and promote parity between languages are all welcome, but requiring companies to adopt certain technologies is not an effective way to achieve those ends.

## INTERNATIONAL AND MULTILATERAL STANDARDS BODIES, REGULATORY AGENCIES, AND OTHERS SHOULD CONVENE MULTI-STAKEHOLDER DISCUSSIONS ABOUT STANDARDS AND GUARDRAILS FOR THE DEVELOPMENT AND USE OF LARGE LANGUAGE MODELS.

The norms around when and how multilingual language models should be deployed are very much in flux. Those norms so far have mostly been established implicitly by technology companies in the ways they build and deploy these models, but trends in these norms may be at odds with the public interest. For instance, OpenAI revealed some information about the training data they used for GPT-3 but almost nothing about GPT-4; Open AI co-founder Ilya Sutskever described having shared information about GPT-3's training data as "just not wise" and something the company would unlikely do again (Vincent, 2023b).

Companies should not have a monopoly on the norms around language models. Governmental and nongovernmental convening bodies need to organize and push back to establish counter-norms that better serve the public's interests. This field is early on enough that these bodies should discuss what positive outcomes even look like. Users affected by the deployment of large language models need to be at the table for those conversations. Government agencies and multilateral organizations (e.g. the Internet Engineering Task Force, United Nations) can play a coordinating role to get together the relevant stakeholders to come up with such standards.

There are also larger questions to reckon with when it comes to the use of large language models in non-English contexts. At once, companies are increasingly deploying multilingual language models to bridge the gap between the functionality in English and other languages across a myriad of tasks, such as harmful content detection, sentiment analysis, and content scanning. However, as we show in this paper, these multilingual systems are relatively new and perform inconsistently across languages.

If deployed prematurely and without guardrails, these models pose real risks to individuals around the world and in particular their ability to express themselves freely. These risks have the potential to compound existing challenges in the information environment for individuals in Western democracies where there are real vacuums of available information in languages other than English and in countries in the Global South where there are already real threats to the free expression and exchange of information posed by majoritarian and institutional powers (Golebiewski & boyd, 2018). Alternatively, companies may decide to only roll out systems that have been fine-tuned for English and wait until there is enough data and tooling available for non-English language tools — something that will take an enormous amount of financial investment, time, effort, and rare consensus — further entrenching the digital divide and Anglocentrism present online. Both scenarios are lose-lose for all speakers on the web. This is a wicked problem and the current incentives are at play to build bigger models, and with more languages. Multi-stakeholder bodies are much better positioned than companies to determine when the risks associated with building larger, more multilingual language models are worth taking.

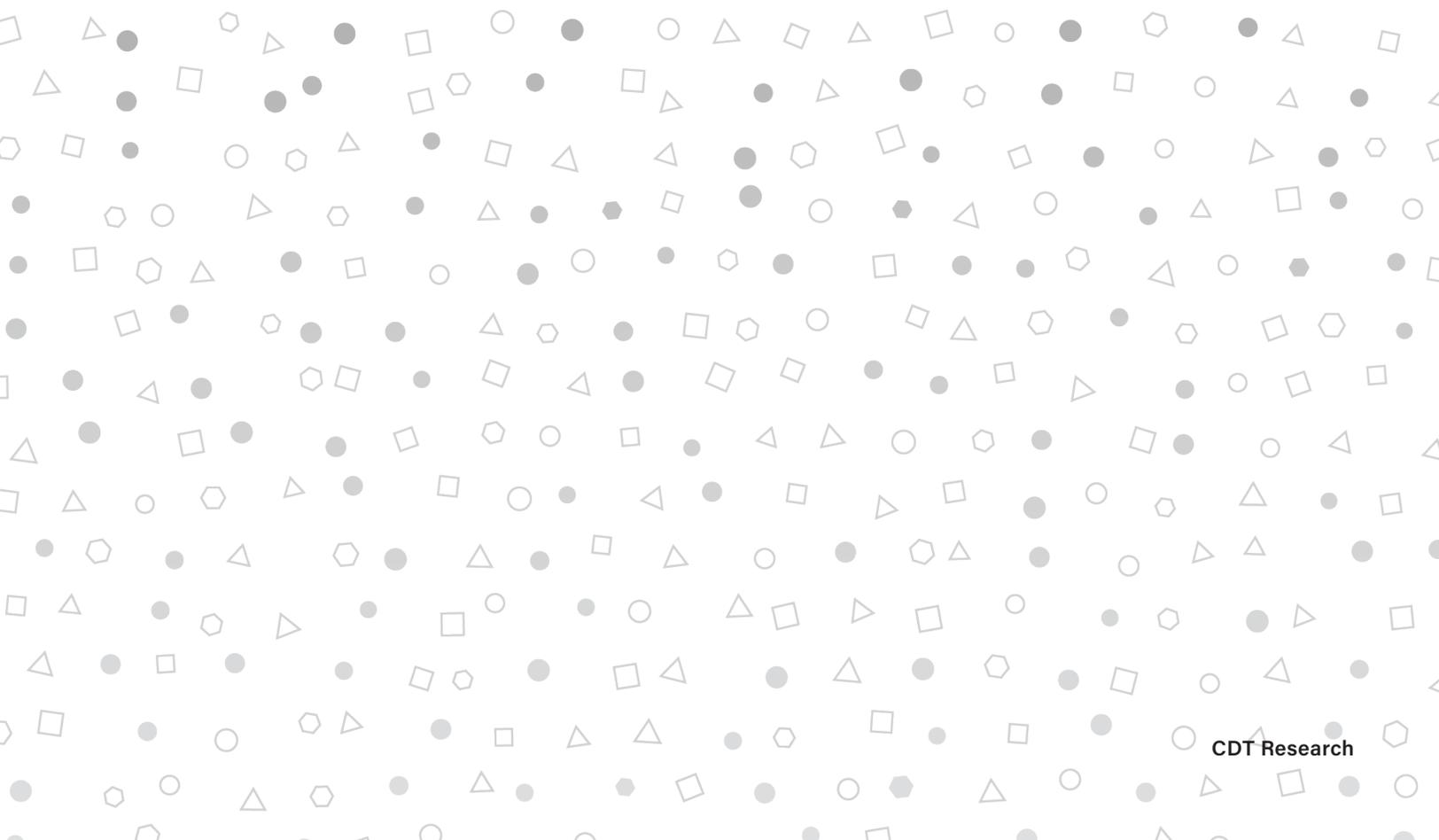

# Works Cited

Abid, A., Farooqi, M., & Zou, J. (2021). Large language models associate Muslims with violence. *Nature Machine Intelligence*, 3(6), Article 6. [perma.cc/HK4B-3AAQ]

ACL. (2021, August 3). *ACL 2022 Theme Track: "Language Diversity: from Low-Resource to Endangered Languages."* ACL. [perma.cc/F2YW-QZBP]

*ACL Rolling Review Dashboard*. (2022). Papers Mentioning >0 Languages. [perma.cc/EQU9-5CWQ]

Agerri, R., Vicente, I. S., Campos, J. A., Barrena, A., Saralegi, X., Soroa, A., & Agirre, E. (2020). Give your Text Representation Models some Love: The Case for Basque. *Proceedings of the 12th Conference on Language Resources and Evaluation*, 4781–4788. [perma.cc/R2DA-GGQZ]

Alyafeai, Z., & Al-Shaibani, M. (2020). ARBML: Democratizing Arabic Natural Language Processing Tools. *Proceedings of Second Workshop for NLP Open Source Software (NLP-OSS)*, 8–13. [perma.cc/4TFY-E9EJ]

Amer, M. (2022, July 13). *Large Language Models and Where to Use Them: Part 2*. Cohere. [perma.cc/CRT5-HDX8]

AmericasNLP. (2022, December 7). *Second Workshop on NLP for Indigenous Languages of the Americas (AmericasNLP)*. [perma.cc/SC88-9WGF]

Amrute, S., Singh, R., & Guzmán, R. L. (2022). *A Primer on AI in/from the Majority World*. Data & Society. [perma.cc/SR8B-J2L9]

Antoun, W., Baly, F., & Hajj, H. (2020). AraBERT: Transformer-based Model for Arabic Language Understanding. *Proceedings of the 4th Workshop on Open-Source Arabic Corpora and Processing Tools, with a Shared Task on Offensive Language Detection*, 9–15. [perma.cc/X5VJ-JKXQ]

Artetxe, M., Labaka, G., & Agirre, E. (2020). Translation Artifacts in Cross-lingual Transfer Learning. *Proceedings of the 2020 Conference on Empirical Methods in Natural Language Processing (EMNLP)*, 7674–7684. [perma.cc/MZY5-DL83]

Artetxe, M., Ruder, S., & Yogatama, D. (2020). On the Cross-lingual Transferability of Monolingual Representations. *Proceedings of the 58th Annual Meeting of the Association for Computational Linguistics*, 4623–4637. [perma.cc/7WMN-5QPR]

Artetxe, M., & Schwenk, H. (2019). Massively Multilingual Sentence Embeddings for Zero-Shot Cross-Lingual Transfer and Beyond. *Transactions of the Association for Computational Linguistics*, 7, 597–610. [perma.cc/LB6R-GH9K]

Article 19. (2022). *Bridging the Gap: Local voices in content moderation. Bosnia and Herzegovina*. [perma.cc/ASU5-ST4N]

Avaaz. (2019). *Megaphone for Hate: Disinformation and Hate Speech on Facebook During Assam's Citizenship Count*. Avaaz. [perma.cc/5MXS-7P7N]

Basile, V., Maro, M. D., Croce, D., & Passaro, L. (2020, December 17). *EVALITA 2020: Overview of the 7th Evaluation Campaign of Natural Language Processing and Speech Tools for Italian*. Seventh Evaluation Campaign of Natural Language Processing and Speech Tools for Italian, Online. [perma.cc/76EK-EJQ8]

Belloni, M. (2021, December 8). Multilingual message content moderation at scale. *Bumble Tech*. [perma.cc/RL2A-L2BD]

Bender, E. (2019, September 15). *The #BenderRule: On Naming the Languages We Study and Why It Matters*. The Gradient. [perma.cc/J3ZM-A5UP]

Bender, E., Gebru, T., McMillan-Major, A., & Shmitchell, S. (2021). On the Dangers of Stochastic Parrots: Can Language Models Be Too Big? 🦜 *Proceedings of the 2021 ACM Conference on Fairness, Accountability, and Transparency*, 610–623. [perma.cc/3KLC-TBUY]

Bender, E., & Koller, A. (2020). Climbing towards NLU: On Meaning, Form, and Understanding in the Age of Data. *Proceedings of the 58th Annual Meeting of the Association for Computational Linguistics*, 5185–5198. [perma.cc/TN3W-5NTC]

Bergman, A., & Diab, M. (2022). Towards Responsible Natural Language Annotation for the Varieties of Arabic. *Findings of the Association for Computational Linguistics: ACL 2022*, 364–371. [perma.cc/Q37M-8F2Y]

BigScience Workshop, Scao, T. L., Fan, A., Akiki, C., Pavlick, E., Ilić, S., Hesslow, D., Castagné, R., Luccioni, A. S., Yvon, F., Gallé, M., Tow, J., Rush, A. M., Biderman, S., Webson, A., Ammanamanchi, P. S., Wang, T., Sagot, B., Muennighoff, N., … Wolf, T. (2023). *BLOOM: A 176B-Parameter Open-Access Multilingual Language Model* (arXiv:2211.05100). arXiv. [perma.cc/2K4Z-F5U7]

Birhane, A., Kalluri, P., Card, D., Agnew, W., Dotan, R., & Bao, M. (2022). The Values Encoded in Machine Learning Research. *2022 ACM Conference on Fairness, Accountability, and Transparency*, 173–184. [perma.cc/9GNB-JHQ5]

Birhane, A., & Prabhu, V. U. (2021). Large image datasets: A pyrrhic win for computer vision? *2021 IEEE Winter Conference on Applications of Computer Vision*, 1536–1546. [perma.cc/Q8LP-THYK]

Bizzoni, Y., Juzek, T. S., España-Bonet, C., Dutta Chowdhury, K., van Genabith, J., & Teich, E. (2020). How Human is Machine Translationese? Comparing Human and Machine Translations of Text and Speech. *Proceedings of the 17th International Conference on Spoken Language Translation*, 280–290. [perma.cc/4DTZ-DVKC]

Bommasani, R., Hudson, D. A., Adeli, E., Altman, R., Arora, S., von Arx, S., Bernstein, M. S., Bohg, J., Bosselut, A., Brunskill, E., Brynjolfsson, E., Buch, S., Card, D., Castellon, R., Chatterji, N., Chen, A., Creel, K., Davis, J. Q., Demszky, D., … Liang, P. (2021). *On the Opportunities and Risks of Foundation Models. Stanford Center for Research on Foundation Models*. [perma.cc/3TKJ-UM2F]

Brown, T., Mann, B., Ryder, N., Subbiah, M., Kaplan, J. D., Dhariwal, P., Neelakantan, A., Shyam, P., Sastry, G., Askell, A., Agarwal, S., Herbert-Voss, A., Krueger, G., Henighan, T., Child, R., Ramesh, A., Ziegler, D., Wu, J., Winter, C., … Amodei, D. (2020). Language Models are Few-Shot Learners. *Advances in Neural Information Processing Systems*, 33, 1877–1901. [perma.cc/7EES-WDAB]

Cahyawijaya, S., Lovenia, H., Aji, A. F., Winata, G. I., Wilie, B., Mahendra, R., Wibisono, C., Romadhony, A., Vincentio, K., Koto, F., Santoso, J., Moeljadi, D., Wirawan, C., Hudi, F., Parmonangan, I. H., Alfina, I., Wicaksono, M. S., Putra, I. F., Rahmadani, S., … Purwarianti, A. (2022). *NusaCrowd: Open Source Initiative for Indonesian NLP Resources* (arXiv:2212.09648). arXiv. [perma.cc/UQ3Y-4LKW]

Callahan, E. S., & Herring, S. C. (2011). Cultural bias in Wikipedia content on famous persons. *Journal of the American Society for Information Science and Technology*, 62(10), 1899–1915. [perma.cc/2S8K-YEJK]

Carlini, N., Ippolito, D., Jagielski, M., Lee, K., Tramer, F., & Zhang, C. (2023, February 1). *Quantifying Memorization Across Neural Language Models*. The Eleventh International Conference on Learning Representations. [perma.cc/678U-9PAQ]

Carlini, N., Tramer, F., Wallace, E., Jagielski, M., Herbert-Voss, A., Lee, K., Roberts, A., Brown, T., Song, D., Erlingsson, U., Oprea, A., & Raffel, C. (2021). *Extracting Training Data from Large Language Models* (arXiv:2012.07805). arXiv. [perma.cc/58MA-VWRZ]

Caswell, I., Breiner, T., van Esch, D., & Bapna, A. (2020). Language ID in the Wild: Unexpected Challenges on the Path to a Thousand-Language Web Text Corpus. *Proceedings of the 28th International Conference on Computational Linguistics*, 6588–6608. [perma.cc/8RFD-DTUK]

Chi, E. A., Hewitt, J., & Manning, C. D. (2020). Finding Universal Grammatical Relations in Multilingual BERT. *Proceedings of the 58th Annual Meeting of the Association for Computational Linguistics*, 5564–5577. [perma.cc/8LNR-VNY9]

Choi, H., Kim, J., Joe, S., Min, S., & Gwon, Y. (2021). *Analyzing Zero-shot Cross-lingual Transfer in Supervised NLP Tasks* (arXiv:2101.10649). arXiv. [perma.cc/NEB9-8THZ]

Chowdhery, A., Narang, S., Devlin, J., Bosma, M., Mishra, G., Roberts, A., Barham, P., Chung, H. W., Sutton, C., Gehrmann, S., Schuh, P., Shi, K., Tsvyashchenko, S., Maynez, J., Rao, A., Barnes, P., Tay, Y., Shazeer, N., Prabhakaran, V., ... Fiedel, N. (2022). *PaLM: Scaling Language Modeling with Pathways*. Google Research. [perma.cc/NZ7N-6GPB]

Christian, J. (2018, July 20). Why Is Google Translate Spitting Out Sinister Religious Prophecies? *Vice*. [perma.cc/8YQU-NUFM]

Conneau, A., Khandelwal, K., Goyal, N., Chaudhary, V., Wenzek, G., Guzmán, F., Grave, E., Ott, M., Zettlemoyer, L., & Stoyanov, V. (2020). Unsupervised Cross-lingual Representation Learning at Scale. *Proceedings of the 58th Annual Meeting of the Association for Computational Linguistics*, 8440–8451. [perma.cc/2MP6-9W3J]

Conneau, A., & Lample, G. (2019). Cross-lingual Language Model Pretraining. *Advances in Neural Information Processing Systems*, 32. [perma.cc/N7EE-JM83]

Conneau, A., Wu, S., Li, H., Zettlemoyer, L., & Stoyanov, V. (2020). Emerging Cross-lingual Structure in Pretrained Language Models. *Proceedings of the 58th Annual Meeting of the Association for Computational Linguistics*, 6022–6034. [perma.cc/3NHR-G7Y4]

Corradi, A. (2017, April 25). *The Linguistic Colonialism of English*. Brown Political Review. [perma.cc/5M3M-9EMN]

Corvey, W. (2014). *Low Resource Languages for Emergent Incidents*. Defense Advanced Research Projects Agency. [perma.cc/4FDR-M3YC]

Crawford, K. (2021). *Atlas of AI: Power, politics, and the planetary costs of artificial intelligence*. Yale University Press.

de Varda, A. G., & Marelli, M. (2023). Data-driven Cross-lingual Syntax: An Agreement Study with Massively Multilingual Models. *Computational Linguistics*, 1–39. [perma.cc/7LQP-EEBQ]

de Vries, W., van Cranenburgh, A., Bisazza, A., Caselli, T., van Noord, G., & Nissim, M. (2019). *BERTje: A Dutch BERT Model* (arXiv:1912.09582). arXiv. [perma.cc/MGU3-WPXR]

Devlin, J., Chang, M.-W., Lee, K., & Toutanova, K. (2019). BERT: Pre-training of Deep Bidirectional Transformers for Language Understanding. *Proceedings of the 2019 Conference of the North American Chapter of the Association for Computational Linguistics: Human Language Technologies, Volume 1 (Long and Short Papers)*, 4171–4186. [perma.cc/E46R-UYDE]

Dodge, J., Sap, M., Marasović, A., Agnew, W., Ilharco, G., Groeneveld, D., Mitchell, M., & Gardner, M. (2021). Documenting Large Webtext Corpora: A Case Study on the Colossal Clean Crawled Corpus. *Proceedings of the 2021 Conference on Empirical Methods in Natural Language Processing*, 1286–1305. [perma.cc/3GC6-UEWJ]

Duarte, N., Llansó, E., & Loup, A. C. (2017). *Mixed Messages? The Limits of Automated Social Media Content Analysis*. Center for Democracy & Technology. [perma.cc/9BRH-5ZZN]

Dulhanty, C., Deglint, J. L., Daya, I. B., & Wong, A. (2019, November 26). *Taking a Stance on Fake News: Towards Automatic Disinformation Assessment via Deep Bidirectional Transformer Language Models for Stance Detection*. NeurIPS 2019, Vancouver. [perma.cc/P5JD-5AD9]

Ebers, M., Poncibò, C., & Zou, M. (Eds.). (2022). *Contracting and Contract Law in the Age of Artificial Intelligence*. Hart Publishing. [perma.cc/G4XR-VYNL]

Eronen, J., Ptaszynski, M., & Masui, F. (2023). Zero-shot cross-lingual transfer language selection using linguistic similarity. *Information Processing & Management*, 60(3), 103250. [perma.cc/S78N-C9MR]

Ethnologue. (2023a). *Assamese*. Ethnologue, Languages of the World. [perma.cc/BE78-H3PN]

Ethnologue. (2023b). *Statistics*. Ethnologue, Languages of the World. [perma.cc/H27U-44TK]

Ganguli, D., Hernandez, D., Lovitt, L., DasSarma, N., Henighan, T., Jones, A., Joseph, N., Kernion, J., Mann, B., Askell, A., Bai, Y., Chen, A., Conerly, T., Drain, D., Elhage, N., Showk, S. E., Fort, S., Hatfield-Dodds, Z., Johnston, S., ... Clark, J. (2022). Predictability and Surprise in Large Generative Models. *2022 ACM Conference on Fairness, Accountability, and Transparency*, 1747–1764. [perma.cc/C8YH-6LMA]

Golebiewski, M., & boyd, danah. (2018). *Data Voids: Where Missing Data Can Easily Be Exploited*. Data & Society. [perma.cc/HE5A-7QTJ]

Góngora, S., Giossa, N., & Chiruzzo, L. (2021). Experiments on a Guarani Corpus of News and Social Media. *Proceedings of the First Workshop on Natural Language Processing for Indigenous Languages of the Americas*, 153–158. [perma.cc/N6S5-4PGN]

Grant-Chapman, H., Laird, E., & Venzke, C. (2021). *Student Activity Monitoring Software Research Insights and Recommendations*. Center for Democracy & Technology. [perma.cc/FY8G-WC2P]

Gröndahl, T., Pajola, L., Juuti, M., Conti, M., & Asokan, N. (2018). All You Need is "Love": Evading Hate Speech Detection. *Proceedings of the 11th ACM Workshop on Artificial Intelligence and Security*, 2–12. [perma.cc/T6P5-FRX5]

Hao, K. (2022, April 22). *A new vision of artificial intelligence for the people*. MIT Technology Review. [perma.cc/54U3-KU5C]

Heikkilä, M. (2022, November 14). *We're getting a better idea of AI's true carbon footprint*. MIT Technology Review. [perma.cc/8PWZ-ESJK]

Hutchinson, B., Prabhakaran, V., Denton, E., Webster, K., Zhong, Y., & Denuyl, S. (2020). Social Biases in NLP Models as Barriers for Persons with Disabilities. *Proceedings of the 58th Annual Meeting of the Association for Computational Linguistics*, 5491–5501. [perma.cc/8FGR-P3FA]

Izsak, P., Berchansky, M., & Levy, O. (2021). How to Train BERT with an Academic Budget. *Proceedings of the 2021 Conference on Empirical Methods in Natural Language Processing*, 10644–10652. [perma.cc/8MPG-W2QE]

Joshi, P., Santy, S., Budhiraja, A., Bali, K., & Choudhury, M. (2020). The State and Fate of Linguistic Diversity and Inclusion in the NLP World. *Proceedings of the 58th Annual Meeting of the Association for Computational Linguistics*, 6282–6293. [perma.cc/82HQ-EH65]

Kaack, L. H., Donti, P. L., Strubell, E., Kamiya, G., Creutzig, F., & Rolnick, D. (2022). Aligning artificial intelligence with climate change mitigation. *Nature Climate Change*, 12(6), Article 6. [perma.cc/7C4S-X2LH]

Khan, M., & Hanna, A. (2023). The Subjects and Stages of AI Dataset Development: A Framework for Dataset Accountability. *Ohio State Technology Law Journal*, 19. [perma.cc/XLG3-AP2J]

Kinchin, N., & Mougouei, D. (2022). What Can Artificial Intelligence Do for Refugee Status Determination? A Proposal for Removing Subjective Fear. *International Journal of Refugee Law*. [perma.cc/3KER-DZ5R]

Koehn, P., & Knowles, R. (2017). Six Challenges for Neural Machine Translation. *Proceedings of the First Workshop on Neural Machine Translation*, 28–39. [perma.cc/9WSQ-HQJY]

Kornai, A. (2013). Digital Language Death. PLOS ONE, 8(10), e77056. [perma.cc/MMZ8-C9VH]

Kreutzer, J., Caswell, I., Wang, L., Wahab, A., van Esch, D., Ulzii-Orshikh, N., Tapo, A., Subramani, N., Sokolov, A., Sikasote, C., Setyawan, M., Sarin, S., Samb, S., Sagot, B., Rivera, C., Rios, A., Papadimitriou, I., Osei, S., Suarez, P. O., ... Adeyemi, M. (2022). Quality at a Glance: An Audit of Web-Crawled Multilingual Datasets. *Transactions of the Association for Computational Linguistics*, 10, 50–72. [perma.cc/YZ7B-Q7PN]

Kublik, V. (n.d.). *EU/US Copyright Law and Implications on ML Training Data*. Valohai. [perma.cc/LD3Z-RVW7]

Kupfer, M., & Muyumba, J. (2022). *Language & Coloniality: Non-Dominant Languages in the Digital Landscape*. Pollicy. [perma.cc/PM8N-Y9YW]

Laurençon, H., Saulnier, L., Wang, T., Akiki, C., Moral, A. V. del, Scao, T. L., Werra, L. V., Mou, C., Ponferrada, E. G., Nguyen, H., Frohberg, J., Šaško, M., Lhoest, Q., McMillan-Major, A., Dupont, G., Biderman, S., Rogers, A., Allal, L. B., Toni, F. D., ... Jernite, Y. (2022, October 31). *The BigScience ROOTS Corpus: A 1.6TB Composite Multilingual Dataset*. Thirty-sixth Conference on Neural Information Processing Systems Datasets and Benchmarks Track. [perma.cc/QS7B-YNYU]

Lauscher, A., Ravishankar, V., Vulić, I., & Glavaš, G. (2020). From Zero to Hero: On the Limitations of Zero-Shot Language Transfer with Multilingual Transformers. *Proceedings of the 2020 Conference on Empirical Methods in Natural Language Processing (EMNLP)*, 4483–4499. [perma.cc/ZJ3R-95JM]

Lees, A., Sorensen, J., & Kivlichan, I. (2020). Jigsaw @ AMI and HaSpeeDe2: Fine-Tuning a Pre-Trained Comment-Domain BERT Model. In V. Basile, D. Croce, M. Maro, & L. C. Passaro (Eds.), *EVALITA Evaluation of NLP and Speech Tools for Italian—December 17th, 2020* (pp. 40–47). Accademia University Press. [perma.cc/9D4M-RSCL]

Lees, A., Tran, V. Q., Tay, Y., Sorensen, J., Gupta, J., Metzler, D., & Vasserman, L. (2022). A New Generation of Perspective API: Efficient Multilingual Character-level Transformers. *Proceedings of the 28th ACM SIGKDD Conference on Knowledge Discovery and Data Mining*, 3197–3207. [perma.cc/5K82-WG8J]

Libovický, J., Rosa, R., & Fraser, A. (2019). *How Language-Neutral is Multilingual BERT?* (arXiv:1911.03310). arXiv. [perma.cc/96RW-WXBL]

Lin, X. V., Mihaylov, T., Artetxe, M., Wang, T., Chen, S., Simig, D., Ott, M., Goyal, N., Bhosale, S., Du, J., Pasunuru, R., Shleifer, S., Koura, P. S., Chaudhary, V., O'Horo, B., Wang, J., Zettlemoyer, L., Kozareva, Z., Diab, M., … Li, X. (2022). Few-shot Learning with Multilingual Generative Language Models. *Proceedings of the 2022 Conference on Empirical Methods in Natural Language Processing*, 9019–9052. [perma.cc/5QY9-97G5]

Lokhov, I. (2021, January 28). *Why are there so many Wikipedia articles in Swedish and Cebuano?* Datawrapper Blog. [perma.cc/WDL2-TF53]

Luccioni, A., & Viviano, J. (2021). What's in the Box? An Analysis of Undesirable Content in the Common Crawl Corpus. *Proceedings of the 59th Annual Meeting of the Association for Computational Linguistics and the 11th International Joint Conference on Natural Language Processing (Volume 2: Short Papers)*, 182–189. [perma.cc/2QQU-NRPB]

Lunden, I. (2023, March 14). Nabla, a digital health startup, launches Copilot, using GPT-3 to turn patient conversations into action. *TechCrunch*. [perma.cc/MK55-SV54]

Martin, G., Mswahili, M. E., Jeong, Y.-S., & Woo, J. (2022). SwahBERT: Language Model of Swahili. *Proceedings of the 2022 Conference of the North American Chapter of the Association for Computational Linguistics: Human Language Technologies*, 303–313. [perma.cc/3ZP6-V6AJ]

Martin, L., Muller, B., Suárez, P. J. O., Dupont, Y., Romary, L., de la Clergerie, É. V., Seddah, D., & Sagot, B. (2020). CamemBERT: A Tasty French Language Model. *Proceedings of the 58th Annual Meeting of the Association for Computational Linguistics*, 7203–7219. [perma.cc/76EU-4LTM]

Masakhane. (n.d.). *Masakhane*. Retrieved December 21, 2022. [perma.cc/A7SA-ALPM]

Maundu, C. (2023, February 21). *How language denies people access to public information*. Nation. [perma.cc/8C4B-JS3Y]

Mazzucato, M. (2014). *The entrepreneurial state: Debunking public vs. private sector myths* (Revised edition). Anthem Press.

Meta AI. (2019, November 7). XLM-R: State-of-the-art cross-lingual understanding through self-supervision. Meta AI. [perma.cc/J55N-4MV5]

Micallef, K., Gatt, A., Tanti, M., van der Plas, L., & Borg, C. (2022). Pre-training Data Quality and Quantity for a Low-Resource Language: New Corpus and BERT Models for Maltese. *Proceedings of the Third Workshop on Deep Learning for Low-Resource Natural Language Processing*, 90–101. [perma.cc/QY8V-9Q6H]

Mikolov, T., Chen, K., Corrado, G., & Dean, J. (2013, September 6). *Efficient Estimation of Word Representations in Vector Space*. International Conference on Learning Representations. [perma.cc/T869-PDX4]

Muller, B., Anastasopoulos, A., Sagot, B., & Seddah, D. (2021). When Being Unseen from mBERT is just the Beginning: Handling New Languages With Multilingual Language Models. *Proceedings of the 2021 Conference of the North American Chapter of the Association for Computational Linguistics: Human Language Technologies*, 448–462. [perma.cc/J5MH-QDW3]

Nadkarni, P. M., Ohno-Machado, L., & Chapman, W. W. (2011). Natural language processing: An introduction. *Journal of the American Medical Informatics Association : JAMIA*, 18(5), 544–551. [perma.cc/72PK-UGK9]

Nekoto, W., Marivate, V., Matsila, T., Fasubaa, T., Fagbohungbe, T., Akinola, S. O., Muhammad, S., Kabongo Kabenamualu, S., Osei, S., Sackey, F., Niyongabo, R. A., Macharm, R., Ogayo, P., Ahia, O., Berhe, M. M., Adeyemi, M., Mokgesi-Selinga, M., Okegbemi, L., Martinus, L., ... Bashir, A. (2020). Participatory Research for Low-resourced Machine Translation: A Case Study in African Languages. *Findings of the Association for Computational Linguistics: EMNLP 2020*, 2144–2160. [perma.cc/5BVM-LUMM]

Nguer, E. M., Lo, A., Dione, C. M. B., Ba, S. O., & Lo, M. (2020). SENCORPUS: A French-Wolof Parallel Corpus. *Proceedings of the Twelfth Language Resources and Evaluation Conference*, 2803–2811. [perma.cc/NBE7-QCZW]

Nguyen, T. (2020, November 27). *Why fake news is so hard to combat in Asian American communities*. Vox. [perma.cc/45GF-UUEC]

Nicholas, G. (2020). Explaining Algorithmic Decisions. *Georgetown Law Technology Review*, 4(711), 20. [perma.cc/UD7D-HF6F]

Nicholas, G. (2022). Shedding Light on Shadowbanning. *Center for Democracy & Technology*. [perma.cc/D2TS-Y92D]

Nkemelu, D., Shah, H., Essa, I., & Best, M. L. (2023). *Tackling Hate Speech in Low-resource Languages with Context Experts*. International Conference on Information & Communication Technologies and Development, Washington, USA. [perma.cc/5QK7-GTMR]

NLLB Team, Costa-jussà, M. R., Cross, J., Çelebi, O., Elbayad, M., Heafield, K., Heffernan, K., Kalbassi, E., Lam, J., Licht, D., Maillard, J., Sun, A., Wang, S., Wenzek, G., Youngblood, A., Akula, B., Barrault, L., Gonzalez, G. M., Hansanti, P., ... Wang, J. (2022). *No Language Left Behind: Scaling Human-Centered Machine Translation* (arXiv:2207.04672). arXiv. [perma.cc/LZH5-DMUA]

Okerlund, J., Klasky, E., Middha, A., Kim, S., Rosenfeld, H., Kleinman, M., & Parthasarathy, S. (2022). *What's in the Chatterbox? Large Language Models, Why They Matter, and What We Should Do About Them*. University of Michigan. [perma.cc/8SXE-RSYE]

OpenAI. (2023). *GPT-4 Technical Report* (arXiv:2303.08774). arXiv. [perma.cc/6ACB-LZYC]

Orife, I., Kreutzer, J., Sibanda, B., Whitenack, D., Siminyu, K., Martinus, L., Ali, J. T., Abbott, J., Marivate, V., Kabongo, S., Meressa, M., Murhabazi, E., Ahia, O., van Biljon, E., Ramkilowan, A., Akinfaderin, A., Öktem, A., Akin, W., Kioko, G., ... Bashir, A. (2020). *Masakhane—Machine Translation For Africa* (arXiv:2003.11529). arXiv. [perma.cc/84Z4-S7AZ]

Patel, F., Levinson-Waldman, R., Koreh, R., & DenUyl, S. (2020). *Social Media Monitoring*. Brennan Center for Justice. [perma.cc/N5LF-ZKP2]

Phillipson, R. (1992). *Linguistic Imperialism*. Oxford University Press.

Pires, T., Schlinger, E., & Garrette, D. (2019). How Multilingual is Multilingual BERT? *Proceedings of the 57th Annual Meeting of the Association for Computational Linguistics*, 4996–5001. [perma.cc/4DPF-LWWX]

Png, M.-T. (2022). At the Tensions of South and North: Critical Roles of Global South Stakeholders in AI Governance. *2022 ACM Conference on Fairness, Accountability, and Transparency*, 1434–1445. [perma.cc/Z7HD-3T4A]

Polignano, M., Basile, P., Degemmis, M., Semeraro, G., & Basile, V. (2019). *AlBERTo: Italian BERT Language Understanding Model for NLP Challenging Tasks Based on Tweets*. Sixth Italian Conference on Computational Linguistics, Bari, Italy. [perma.cc/RBY9-4JHJ]

Prabhakaran, V., Mitchell, M., Gebru, T., & Gabriel, I. (2022). *A Human Rights-Based Approach to Responsible AI* (arXiv:2210.02667). arXiv. [perma.cc/R97H-WQSK]

Prates, M., Avelar, P., & Lamb, L. (2020). Assessing gender bias in machine translation: A case study with Google Translate. *Neural Computing and Applications*, 32. [perma.cc/CGK2-NMU2]

Pym, A., Ayvazyan, N., & Prioleau, J. M. (2022). Should raw machine translation be used for public-health information? Suggestions for a multilingual communication policy in Catalonia. *Just. Journal of Language Rights & Minorities, Revista de Drets Lingüístics i Minories*, 1(1–2), 71–99. [perma.cc/HSA8-TB3F]

Raji, D., Denton, E., Bender, E. M., Hanna, A., & Paullada, A. (2021). AI and the Everything in the Whole Wide World Benchmark. *Proceedings of the Neural Information Processing Systems Track on Datasets and Benchmarks*, 1. [perma.cc/EX84-X9BQ]

Razeghi, Y., Logan IV, R. L., Gardner, M., & Singh, S. (2022). Impact of Pretraining Term Frequencies on Few-Shot Numerical Reasoning. *Findings of the Association for Computational Linguistics: EMNLP 2022*, 840–854. [perma.cc/SMG9-BSKV]

Reid, M., & Artetxe, M. (2022). *On the Role of Parallel Data in Cross-lingual Transfer Learning* (arXiv:2212.10173). arXiv. [perma.cc/83GW-CVXX]

Richter, F. (n.d.). *English Is the Internet's Universal Language*. Statista Infographics. Retrieved December 14, 2022, from [perma.cc/WW7B-7X37]

Rionda, V. P. S., & Mejia, J. C. U. (2021). *PretorIA y la automatización del procesamiento de causas de derechos humanos*. Derechos Digitales and Dejusticia. [perma.cc/65MQ-X484]

Rowe, J. (2022, March 2). *Marginalised languages and the content moderation challenge*. Global Partners Digital. [perma.cc/GU4K-5HBE]

Ruder, S., Constant, N., Botha, J., Siddhant, A., Firat, O., Fu, J., Liu, P., Hu, J., Garrette, D., Neubig, G., & Johnson, M. (2021). XTREME-R: Towards More Challenging and Nuanced Multilingual Evaluation. *Proceedings of the 2021 Conference on Empirical Methods in Natural Language Processing*, 10215–10245. [perma.cc/W4TJ-SGTB]

Santa Clara Principles. (2021). *Santa Clara Principles on Transparency and Accountability in Content Moderation*. Santa Clara Principles. [perma.cc/T623-AVW6]

Santy, S., Kummerfeld, J., & Rubio, H. (2023). *Languages mentioned in Paper Abstracts*. ACL Rolling Review. [perma.cc/EQU9-5CWQ]

Savoldi, B., Gaido, M., Bentivogli, L., Negri, M., & Turchi, M. (2021). Gender Bias in Machine Translation. *Transactions of the Association for Computational Linguistics*, 9, 845–874. [perma.cc/9K3F-5VBZ]

Schwenk, H. (2019, January 22). *LASER natural language processing toolkit—Engineering at Meta*. Meta AI. [perma.cc/46JG-AZ4T]

Sengupta, P. B., Claudia Pozo, Anasuya. (2022, March 31). Does the internet speak your language? Launching the first-ever State of the Internet's Languages report. *Whose Knowledge?* [https://perma.cc/9KCX-M863]

Sharir, O., Peleg, B., & Shoham, Y. (2020). *The Cost of Training NLP Models: A Concise Overview* (arXiv:2004.08900). arXiv. [perma.cc/8KVV-C6P2]

Shenkman, C., Thakur, D., & Llansó, E. (2021). *Do You See What I See? Capabilities and Limits of Automated Multimedia Content Analysis*. Center for Democracy & Technology. [perma.cc/W85T-HQQF]

Srivastava, A., Rastogi, A., Rao, A., Shoeb, A. A. M., Abid, A., Fisch, A., Brown, A. R., Santoro, A., Gupta, A., Garriga-Alonso, A., Kluska, A., Lewkowycz, A., Agarwal, A., Power, A., Ray, A., Warstadt, A., Kocurek, A. W., Safaya, A., Tazarv, A., … Wu, Z. (2022). *Beyond the Imitation Game: Quantifying and extrapolating the capabilities of language models* (arXiv:2206.04615). arXiv. [perma.cc/278S-ZJV9]

Stadthagen-Gonzalez, H., Imbault, C., Pérez Sánchez, M. A., & Brysbaert, M. (2017). Norms of valence and arousal for 14,031 Spanish words. *Behavior Research Methods*, 49(1), 111–123. [perma.cc/7FWX-Z3JD]

Strubell, E., Ganesh, A., & McCallum, A. (2019). Energy and Policy Considerations for Deep Learning in NLP. *Proceedings of the 57th Annual Meeting of the Association for Computational Linguistics*, 3645–3650. [perma.cc/9P4Y-J4HT]

Talat, Z., Névéol, A., Biderman, S., Clinciu, M., Dey, M., Longpre, S., Luccioni, S., Masoud, M., Mitchell, M., Radev, D., Sharma, S., Subramonian, A., Tae, J., Tan, S., Tunuguntla, D., & Wal, O. van der. (2022). You reap what you sow: On the Challenges of Bias Evaluation Under Multilingual Settings. *Proceedings of BigScience Episode #5*, 26–41. [perma.cc/3ECR-4E7U]

Tattle. (n.d.). *Uli*. [perma.cc/4AB2-D4GX]

Tay, Y., Tran, V. Q., Ruder, S., Gupta, J., Chung, H. W., Bahri, D., Qin, Z., Baumgartner, S., Yu, C., & Metzler, D. (2022, February 23). *Charformer: Fast Character Transformers via Gradient-based Subword Tokenization*. International Conference on Learning Representations 2022. [perma.cc/YRL4-E7DT]

Teich, E. (2003). Cross-Linguistic Variation in System and Text: A Methodology for the Investigation of Translations and Comparable Texts. In *Cross-Linguistic Variation in System and Text*. De Gruyter Mouton. [perma.cc/L8A8-RH8B]

Torbati, Y. (2019, September 26). *Google Says Google Translate Can't Replace Human Translators. Immigration Officials Have Used It to Vet Refugees*. *ProPublica*. [perma.cc/ZUN6-LHA5]

Toyama, K. (2015). *Geek heresy: Rescuing social change from the cult of technology*. PublicAffairs.

Tsvetkov, Y., Sitaram, S., Faruqui, M., Lample, G., Littell, P., Mortensen, D., Black, A. W., Levin, L., & Dyer, C. (2016). Polyglot Neural Language Models: A Case Study in Cross-Lingual Phonetic Representation Learning. *Proceedings of the 2016 Conference of the North American Chapter of the Association for Computational Linguistics: Human Language Technologies*, 1357–1366. [perma.cc/4RES-KFNM]

United Nations Human Rights Office of the High Commissioner. (n.d.). *Economic, social and cultural rights*. OHCHR. [perma.cc/Y6MK-SZZ4]

Vallee, H. Q. la, & Duarte, N. (2019). Algorithmic Systems in Education: Incorporating Equity and Fairness When Using Student Data. *Center for Democracy and Technology*. [perma.cc/CC89-ZVNV]

Vaswani, A., Shazeer, N., Parmar, N., Uszkoreit, J., Jones, L., Gomez, A. N., Kaiser, L., & Polosukhin, I. (2017, December 5). *Attention Is All You Need. Advances in Neural Information Processing Systems*. [perma.cc/2ZDX-Z796]

Vincent, J. (2022, November 2). *Google plans giant AI language model supporting world's 1,000 most spoken languages*. The Verge. [perma.cc/3Y48-X7WV]

Vincent, J. (2023a, January 17). *Getty Images is suing the creators of AI art tool Stable Diffusion for scraping its content*. The Verge. [perma.cc/4CXS-3WNN]

Vincent, J. (2023b, March 15). *OpenAI co-founder on company's past approach to openly sharing research: "We were wrong."* The Verge. [perma.cc/DPL6-4PD2]

Vitulli, M. A. (2018). Writing Women in Mathematics Into Wikipedia. *Notices of the American Mathematical Society*, 65(03), 330–334. [perma.cc/X73F-AZPM]

Volansky, V., Ordan, N., & Wintner, S. (2015). On the features of translationese. *Digital Scholarship in the Humanities*, 30(1), 98–118. [perma.cc/7F8S-3YXK]

Wang, C., Cho, K., & Gu, J. (2020). Neural Machine Translation with Byte-Level Subwords. *Proceedings of the AAAI Conference on Artificial Intelligence*, 34(05), Article 05. [perma.cc/5DL7-XSSP]

Wang, Z., K, K., Mayhew, S., & Roth, D. (2020). Extending Multilingual BERT to Low-Resource Languages. *Findings of the Association for Computational Linguistics: EMNLP 2020*, 2649–2656. [perma.cc/ZNC8-C9E7]

Williams, A., Miceli, M., & Gebru, T. (2022). The Exploited Labor Behind Artificial Intelligence. *Noēma*. [perma.cc/GE8H-7SUN]

Wu, S., & Dredze, M. (2019). Beto, Bentz, Becas: The Surprising Cross-Lingual Effectiveness of BERT. *Proceedings of the 2019 Conference on Empirical Methods in Natural Language Processing and the 9th International Joint Conference on Natural Language Processing (EMNLP-IJCNLP)*, 833–844. [perma.cc/EJ3G-MFYN]

Wu, S., & Dredze, M. (2020). Are All Languages Created Equal in Multilingual BERT? *Proceedings of the 5th Workshop on Representation Learning for NLP*, 120–130. [perma.cc/5E6X-NNAA]

Yu, S., Sun, Q., Zhang, H., & Jiang, J. (2022). Translate-Train Embracing Translationese Artifacts. *Proceedings of the 60th Annual Meeting of the Association for Computational Linguistics (Volume 2: Short Papers)*, 362–370. [perma.cc/7F8C-EYM6]

Zhang, S., Frey, B., & Bansal, M. (2022, April 25). How can NLP Help Revitalize Endangered Languages? A Case Study and Roadmap for the Cherokee Language. *Proceedings of the 60th Annual Meeting of the Association for Computational Linguistics (Volume 1: Long Papers)*. ACL 2022, Dublin, Ireland. [perma.cc/2XF2-2GDC]

Zhang, Y., Warstadt, A., Li, X., & Bowman, S. R. (2021). When Do You Need Billions of Words of Pretraining Data? *Proceedings of the 59th Annual Meeting of the Association for Computational Linguistics and the 11th International Joint Conference on Natural Language Processing (Volume 1: Long Papers)*, 1112–1125. [perma.cc/43ZK-2ZXC]

cdt.org

cdt.org/contact

**Center for Democracy & Technology**
1401 K Street NW, Suite 200
Washington, D.C. 20005

202-637-9800

@CenDemTech

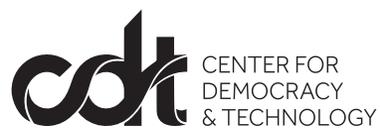